\documentclass[runningheads]{llncs}
\usepackage{graphicx}

\usepackage{tikz}
\usepackage{comment}
\usepackage{amsmath,amssymb} 
\usepackage{color}
\usepackage{sidecap}
\usepackage{algorithm}
\usepackage{algpseudocode}
\usepackage[symbol]{footmisc}

\usepackage[accsupp]{axessibility}  

\usepackage{tabularx}
\usepackage{xcolor}
\usepackage{multirow}
\usepackage{multicol,lipsum}
\usepackage{booktabs}
\usepackage[export]{adjustbox}
\usepackage{colortbl}
\usepackage{makecell}
\usepackage[pagebackref,breaklinks,colorlinks]{hyperref}

\def\I{{\bf I}}

\def\z{{\bf z}}

\def\0{{\bf 0}}
\def\1{{\bf 1}}

\def\muu{\mbox{\boldmath$\mu$\unboldmath}}

\definecolor{red}{rgb}{0.95,0.4,0.4}
\definecolor{purered}{rgb}{1,0,0}
\definecolor{blue}{rgb}{0.4,0.4,0.95}
\definecolor{darkblue}{rgb}{0,0,0.8}
\definecolor{darkred}{rgb}{1,0,0}
\definecolor{darkgreen}{rgb}{0,0.5,0}
\definecolor{grey}{rgb}{0.6,0.6,0.6}
\definecolor{col1}{RGB}{232, 161, 148}
\definecolor{col2}{RGB}{148, 187, 232}
\definecolor{lightgrey}{rgb}{0.85,0.85,0.85}
\definecolor{lightlightgrey}{rgb}{0.9,0.9,0.9}
\definecolor{verylightBG}{rgb}{0.9,0.99,0.99}
\definecolor{darkgreen}{rgb}{0.3, 0.75, 0.3}

\graphicspath{{./figure/}}   

\usepackage[capitalize]{cleveref}
\crefname{section}{Sec.}{Secs.}
\Crefname{section}{Section}{Sections}
\Crefname{table}{Table}{Tables}
\crefname{table}{Tab.}{Tabs.}

\begin{document}
\pagestyle{headings}
\mainmatter
\def\ECCVSubNumber{1448}  

\title{Multimodal Object Detection via Probabilistic Ensembling} 

\titlerunning{Multimodal Object Detection via Probabilistic Ensembling}

\authorrunning{Chen et al.}

\author{Yi-Ting Chen\thanks{Equal contribution. Most of the work was done when authors were with CMU.}\inst{1},
Jinghao Shi\textsuperscript{$\ast$}\inst{2},
Zelin Ye\textsuperscript{$\ast$}\inst{2},
Christoph Mertz\inst{2},   \\
Deva Ramanan\thanks{Equal supervision.}\inst{2,3},
Shu Kong\textsuperscript{$\dagger$}\inst{2,4}
}

\institute{
University of Maryland, College Park \and
Carnegie Mellon University \and 
Argo AI  \and 
Texas A\&M University \\
\texttt{\small \\
ytchen@umd.edu, \{jinghaos, zeliny, cmertz\}@andrew.cmu.edu, \\
deva@cs.cmu.edu,  shu@tamu.edu}\\
\href{https://github.com/Jamie725/RGBT-detection}{open-source code in Github}
\vspace{-3mm}
}

\maketitle

\begin{abstract}
Object detection with multimodal inputs can improve many safety-critical systems such as autonomous vehicles (AVs). Motivated by AVs that operate in both day and night, we study multimodal object detection with RGB and thermal cameras, since the latter provides much stronger object signatures under poor illumination. We explore strategies for fusing information from different modalities. Our key contribution is a probabilistic ensembling technique, {\bf ProbEn}, a simple non-learned method that fuses together detections from multi-modalities. We derive ProbEn from Bayes' rule and first principles that assume conditional independence across modalities. Through probabilistic marginalization, ProbEn elegantly handles missing modalities when detectors do not fire on the same object. Importantly, ProbEn also notably improves multimodal detection even when the conditional independence assumption does not hold, e.g., fusing outputs from other fusion methods  (both off-the-shelf and  trained in-house). We validate ProbEn on two benchmarks containing both aligned (KAIST) and unaligned (FLIR) multimodal images, showing that ProbEn outperforms prior work by more than {\bf 13\%} in relative performance!

\keywords{Object Detection \and Multimodal Detection \and Infrared \and Thermal \and Probabilistic Model \and Ensembling \and Multimodal Fusion \and Uncertainty}
\end{abstract}

\section{Introduction}
Object detection is a canonical computer vision problem that has been greatly advanced by the end-to-end training of deep neural detectors~\cite{ren2015faster,he2017mask}.
Such detectors are widely adopted in various safety-critical systems such as autonomous vehicles (AVs)~\cite{Geiger2012CVPR,caesar2020nuscenes}.
Motivated by AVs that operate in both day and night, we study multimodal object detection with RGB and thermal cameras, since the latter can provide much stronger object signatures under poor illumination~\cite{hwang2015multispectral,xu2017learning,li2018multispectral,devaguptapu2019borrow,zhang2019weakly,bertinitask}. 

\begin{figure}[t]
\centering
\includegraphics[width=0.9\linewidth]{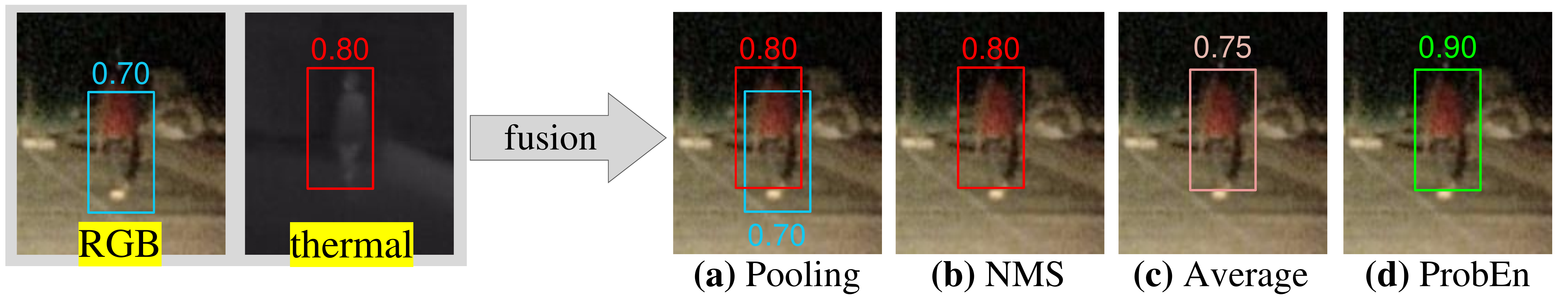}
\vspace{-4mm}
\caption{\small
{\bf Multimodal detection via ensembling single-modal detectors}.
{\bf (a)} A naive approach is to pool detections from each modality, but this will result in multiple detections that overlap the same object. 
({\bf b)} To remedy this, one can apply non-maximal suppression (NMS) to suppress overlapping detections from different modalities, which always returns the higher (maximal) scoring detection. Though quite simple,  NMS is an effective fusion strategy that has \emph{not} been previously proposed as such. However, NMS fails to incorporate cues from the lower-scoring modality. 
{\bf (c)} A natural strategy for doing so might average scores of overlapping detections (instead of suppressing the weaker ones)~\cite{li2019illumination,liu2016multispectral}. However, this must decrease the reported score compared to NMS. 
Intuitively, if two modalities agree on a candidate detection, 
one should {\em boost} its score. 
{\bf (d)} To do so, we derive a simple probabilistic ensembling approach, {\bf ProbEn}, to score fusion that increases the score for detections that have strong evidence from multiple modalities. We further extend ProbEn to box fusion in Section~\ref{sec:fusion}. 
Our {\em non-learned} ProbEn significantly outperforms prior work (Table~\ref{tab:SOTA_KAIST}\&\ref{tab:SOTA_FLIR}).
}
\label{fig:teaser}
\vspace{-3mm}
\end{figure}

{\bf Multimodal Data}.
There exists several challenges in multimodal detection. One is the lack of data. While there exists large repositories of annotated single-modal datasets (RGB) and pre-trained models, there exists much less annotated data of other modalities (thermal), and even less annotations of them paired together. One often-ignored aspect is the alignment of the modalities: aligning RGB and thermal images requires special purpose hardware, e.g., a beam-splitter~\cite{hwang2015multispectral} or a specialized rack~\cite{valverde2021there} for spatial alignment, and a GPS clock synchronizer for temporal alignment~\cite{quigley2009ros}.
Fusion on {\em un}aligned RGB-thermal inputs (cf. Fig.~\ref{fig:FLIR-annotation}) remains relatively unexplored. For example, even annotating bounding boxes is cumbersome because separate annotations are required for each modality, increasing overall cost. As a result, many unaligned datasets annotate only one modality (e.g., FLIR~\cite{FLIR}), further complicating multimodal learning.

\begin{figure}[t]
\centering
\begin{minipage}[l]{0.55\textwidth}
    \centering
    \includegraphics[width=1\linewidth]{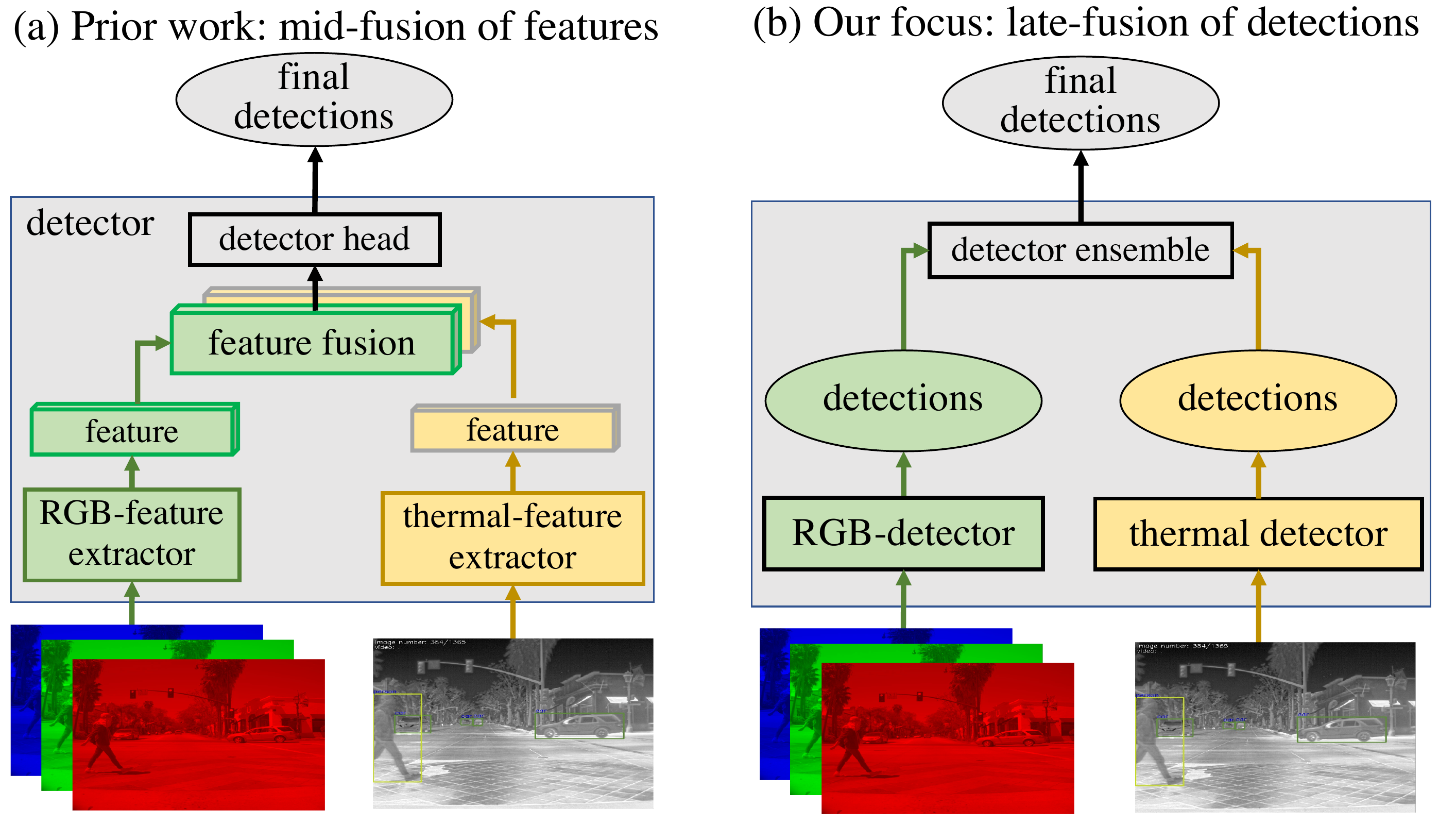}
\end{minipage} \hfill
\begin{minipage}[r]{0.44\textwidth}
\vspace{-3mm}
\caption{\small
    High-level comparisons between mid- and late-fusion. {\bf (a)} Past work primarily focuses on  mid-fusion, e.g., concatenating features computed by single-modal feature extractors. {\bf (b)} We focus on late-fusion via \emph{detector ensemble} that fuses detections from independent detectors, e.g., two single-modal detectors trained with RGB and thermal images respectively.
}
\label{fig:flowchart}
\end{minipage}
\vspace{-6mm}
\end{figure}

{\bf Multimodal Fusion.}
The central question in multimodal detection is {\em how} to fuse information from different modalities.
Previous work has explored strategies for fusion at various stages~\cite{choi2016multi,xu2017learning,li2018multispectral,zhang2019cross,zhang2019weakly,bertinitask}, which are often categorized into early-, mid- and late-fusion.
Early-fusion constructs a four-channel RGB-thermal input~\cite{wagner2016multispectral}, which is then processed by a (typical) deep network. In contrast, mid-fusion keeps RGB and thermal inputs in different streams and then merges their features downstream within the network (Fig.~\ref{fig:flowchart}a)~\cite{wagner2016multispectral,liu2016multispectral,konig2017fully}. The vast majority of past work focuses on architectural design of where and how to merge. 
Our key contribution is the exploration of an extreme variant of {\em very}-late fusion of detectors trained on separate modalities (Fig.~\ref{fig:flowchart}b) through {\em detector ensembling}.
Though conceptually simple, ensembling can be effective because one can learn from single-modal datasets that often dwarf the size of multimodal datasets.
However, ensembling can be practically challenging because different detectors might not fire on the same object. For example, RGB-based detectors often fail to fire in nighttime conditions, implying one needs to deal with ``missing" detections during fusion.

{\bf Probabilistic Ensembling (ProbEn).} 
We derive our very-late fusion approach, ProbEn, from first principles: simply put, 
if single-modal signals are conditionally independent of each other given the true label, the optimal fusion strategy is given by Bayes rule~\cite{pearl2014probabilistic}. ProbEn requires no learning, and so does not require any multimodal data for training. Importantly, ProbEn elegantly handles ``missing" modalities via probabilistic marginalization.
While ProbEn is derived assuming conditional independence, we empirically find that it can be used to fuse outputs that are not strictly independent, by fusing outputs from {\em other} fusion methods (both off-the-shelf and trained in-house).
In this sense, ProbEn is a general technique for ensembling detectors. 
We achieve significant improvements over prior art, both on aligned and unaligned multimodal benchmarks.

{\bf Why ensemble?} 
One may ask why detector ensembling should be regarded as an interesting contribution, 
given that ensembling is a well-studied approach~\cite{freund1996experiments,kittler1998combining,bauer1999empirical,dietterich2000ensemble} that is often viewed as an ``engineering detail'' for improving leaderboard performance~\cite{krizhevsky2012imagenet,huang20201st,guo20192nd}.
Firstly, we show that the precise ensembling technique matters, and prior approaches proposed in the (single-modal) detection literature such as score-averaging~\cite{krizhevsky2012imagenet,dollar2009pedestrian} or max-voting~\cite{xu2014evidential}, are not as effective as ProbEn, particularly when dealing with missing modalities. 
Secondly, to our knowledge, we are the first to propose detector ensembling as a fusion method for multimodal detection. Though quite simple, it is remarkably effective and should be considered a baseline for future research.

\section{Related Work}

{\bf Object Detection and Detector Ensembling.}
State-of-the-art detectors  train deep neural networks on large-scale datasets such as COCO~\cite{lin2014microsoft} and often focus on architectural design \cite{liu2016ssd,redmon2016you,redmon2017yolo9000,ren2015faster}. Crucially, most architectures generate overlapping detections which need to be post-processed with non-maximal suppression (NMS)~\cite{dalal2005histograms,bodla2017soft,solovyev2021weighted}. Overlapping detections could also be generated by detectors tuned for different image crops and scales, which typically make use of ensembling techniques for post-processing their output \cite{akiba2018pfdet,guo20192nd,huang20201st}. Somewhat surprisingly, although detector ensembling and NMS are widely studied in single-modal RGB detection, to the best of our knowledge, they have \emph{not} been used to (very) late-fuse multimodal detections; we find them remarkably effective.

{\bf Multimodal Detection},
particularly with RGB-thermal images, has attracted increasing attention.
The KAIST pedestrian detection dataset~\cite{hwang2015multispectral} is one of the first benchmarks for RGB-thermal detection, fostering
growth of research in this area.
Inspired by the successful RGB-based detectors~\cite{ren2015faster,redmon2016you,liu2016ssd}, current multimodal detectors train deep models with various methods for fusing multimodal signals~\cite{choi2016multi,xu2017learning,li2018multispectral,zhang2019cross,zhang2019weakly,bertinitask,zhou2020improving,zhang2019weakly,kim2021mlpd}. 
Most of these multimodal detection methods work on aligned RGB-thermal images, but it is unclear how they perform on heavily unaligned modalities such as images in Fig.~\ref{fig:FLIR-annotation} taken from FLIR  dataset~\cite{FLIR}.
We study multimodal detection under both aligned and unaligned RGB-thermal scenarios.
{\bf Multimodal fusion} is the central question in multimodal detection.
Compared to early-fusion that simply concatenates RGB and thermal inputs, 
mid-fusion of single-modal features performs better~\cite{wagner2016multispectral}. 
Therefore, most multimodal methods study how to fuse features and focus on designing new network architectures~\cite{wagner2016multispectral,liu2016multispectral,konig2017fully}.
Because RGB-thermal pairs might not be aligned, some methods train an RGB-thermal translation network to synthesize aligned pairs, but this requires annotations in each modality~\cite{devaguptapu2019borrow,munir2020thermal,kieubottom}.
Interestingly, few works explore learning from unaligned data that are annotated only in single modality; we show that mid-fusion architectures can still learn in this setting by acting as an implicit alignment network.
Finally, few fusion architectures explore (very) late fusion of single-modal detections via detector ensembling. Most that do simply take heuristic (weighted) averages of confidence scores~\cite{guan2019fusion,li2018multispectral,zhang2019weakly}.
In contrast, we introduce probabilistic ensembling (ProbEn) for late-fusion, which significantly outperforms prior methods on both aligned and unaligned RGB-thermal data.


\section{Fusion Strategies for Multimodal Detection}
\label{sec:fusion}
We now present multimodal fusion strategies for detection.
We first point out that {\bf single-modal} detectors are viable methods for processing multimodal signals, and so include them as a baseline. 
We also include fusion baselines for {\bf early-fusion}, which concatenates RGB and thermal as a four-channel input, and {\bf mid-fusion}, which concatenates single-modal features inside a network (Fig.~\ref{fig:flowchart}). As a preview of results, we find that mid-fusion is generally the most effective baseline (Table~\ref{tab:ablation_KAIST}). Surprisingly, this holds even for unaligned data that is annotated with a single modality (Fig.~\ref{fig:FLIR-annotation}), indicating that mid-fusion can perform some implicit alignment (Table~\ref{tab:ablation_FLIR}).

We describe strategies for late-fusing detectors from different modalities, or detector ensembling. We begin with a naive approach (Fig.~\ref{fig:teaser}).
Late-fusion needs to fuse scores and boxes; we discuss the latter at the end of this section.

{\bf Naive Pooling}.
The possibly simplest strategy is to naively pool detections from multiple modalities together. This will probably result in multiple detections overlapping the same ground-truth object (Fig.~\ref{fig:teaser}a).

{\bf Non-Maximum Supression (NMS)}. The natural solution for dealing with overlapping detections is NMS, a crucial component in contemporary RGB detectors~\cite{dollar2009pedestrian,zitnick2014edge,hosang2017learning}.
NMS finds bounding box predictions with high spatial overlap and remove the lower-scoring bounding boxes. This can be implemented in a sequential fashion via sorting of predictions by confidence, as depicted by Algorithm~\ref{alg:NMS-for-fusion}, or in a parallel fashion amenable to GPU computation~\cite{bolya2019yolact}. While NMS has been used to ensemble single-modal detectors~\cite{solovyev2021weighted}, it has (surprisingly) \emph{not} been advocated for fusion of {\em multi}-modal detectors. We find it be shockingly effective, outperforming the majority of past work on established benchmarks (Fig.~\ref{tab:SOTA_KAIST}). Specifically, when two detections from two different modalities overlap (e.g., IoU$>$0.5), NMS simply keeps the higher-score detection and suppresses the other (Fig.~\ref{fig:teaser}b). This allows each modality to ``shine" where effective -- thermal detections tend to score high (and so will be selected) when RGB detections perform poorly due to poor illumination conditions. 
That said, rather than selecting one modality at the global image level (e.g., day-time vs. night time), NMS selects one modality at the local bounding box level. However, in some sense, NMS fails to ``fuse" information from multiple modalities together, since each of the final detections are supported by only one modality.

{\bf Average  Fusion}.
To actually fuse multimodal information, a straightforward strategy is to modify NMS to average confidence scores of overlapping detections from different modalities, rather than suppressing the weaker modality. Such an averaging has been proposed in prior work~\cite{xu2014evidential,liu2016multispectral,li2018multispectral}.
However, averaging scores will necessarily {\em decrease} the NMS score which reports the max of an overlapping set of detections (Fig.~\ref{fig:teaser}c). Our experiments demonstrate that averaging produces worse results than NMS and single-modal detectors. Intuitively, if two modalities agree that there exist a detection, fusion should {\em increase} the overall confidence rather than decrease.

{
\setlength{\textfloatsep}{0pt} 
\begin{algorithm}[t]
\small
\caption{Multimodal Fusion by NMS or ProbEn}
\begin{algorithmic}[1]
\State Input: class priors $\pi_k$ for $k\in\{1,\dots,K\}$; the flag of fusion method ({\tt NMS} or {\tt ProbEn});
set $\cal D$: detections from multiple modalities. Each  detection $d=({\bf y},{\bf z}, m) \in \cal D$ contains classification posteriors ${\bf y}$, box coordinates ${\bf z}$ and modality tag $m$.
    \State Initialize set of fused detections $\cal F=\{\}$ 
     \While{${\cal D} \not=\emptyset$}
    \State Find detection $d \in \cal D$ with largest posterior
    \State Find all detections in $\cal D$ that overlap $d$  (e.g., $>$ 0.5 IoU), denoted as ${\cal T} \subseteq {\cal D}$
    \If {{\tt NMS}}
        \State $d' \leftarrow d$ 
    \ElsIf  {{\tt ProbEn}}
        \State Find highest scoring detection in $\cal T$ of each modality,  denoted as ${\cal S} \subseteq {\cal T}$
        \State Compute $d'$ from $\cal S$ by fusing scores ${\bf y}$ 
        with Eq.~\eqref{eq:fuse} and  boxes ${\bf z}$ with Eq.~\eqref{eq:bbox-fusion}
    \EndIf
    \State ${\cal F} \leftarrow {\cal F}+\{d'\}$, \ \ \ \ \ 
    ${\cal D} \leftarrow {\cal D} - {\cal T}$ 
    \EndWhile \\
    \Return   set $\cal F$ of fused detections 
\end{algorithmic}
\label{alg:NMS-for-fusion}
\end{algorithm}
}

{\bf Probabilistic Ensembling (ProbEn)}.
We derive our probabilistic approach for late-fusion of detections by starting with how to fuse detection scores (Algorithm~\ref{alg:NMS-for-fusion}). 
Assume we have an object with label $y$ (e.g., a ``person'') and measured signals from two modalities: $x_1$ (RGB) and $x_2$ (thermal). We write out our formulation for two modalities, but the extension to multiple (evaluated in our experiments) is straightforward. Crucially, we assume measurements are conditionally independent given the object label $y$:
\begin{align}
    p(x_1,x_2|y) = p(x_1|y) p(x_2|y) \label{eq:ind}
\end{align} 
This can also be written as $p(x_1|y) = p(x_1|x_2,y)$, which may be easier to intuit. Given the person label $y$, predict its RGB appearance  $x_1$; if this prediction would not change the given knowledge of the thermal signal $x_2$, then conditional independence holds. 
We wish to infer labels given multimodal measurements: 
\vspace{-4mm}
\begin{align}\small
    p(y|x_1,x_2) = \frac{p(x_1,x_2|y)p(y)}{p(x_1,x_2)}
    \propto p(x_1,x_2|y)p(y) \label{eq:bayes}
\end{align}
By applying the conditional independence assumption from \eqref{eq:ind} to \eqref{eq:bayes}, we have:
\vspace{-4mm}
\begin{align}\small
p(y|x_1,x_2) \propto \   p(x_1|y)p(x_2|y)p(y) 
    &\propto \frac{p(x_1|y)p(y)p(x_2|y)p(y)}{p(y)}\\
    &\propto \frac{p(y|x_1) p(y|x_2)}{p(y)} 
    \label{eq:fuse} 
\end{align}
The above suggests a simple approach to fusion that is provably optimal when single-modal features are conditionally-independent of the true object label:
\begin{enumerate}
    \item Train independent single-modal classifiers that predict the distributions over the label $y$ given each individual feature modality $p(y|x_1)$ and $p(y|x_2)$.
    \item Produce a final score by multiplying the two distributions, dividing by the class prior distribution, and normalizing the final result \eqref{eq:fuse} to sum-to-one.
\end{enumerate}
To obtain the class prior $p(y)$, we can simply  normalize the counts of per-class examples.
Extending ProbEn~\eqref{eq:fuse} to $M$ modalities is simple:
\vspace{-2mm}
\begin{align}\small
p(y|\{x_i\}_{i=1}^M) \propto \frac{\Pi_{i=1}^Mp(y|x_i)}{p(y)^{M-1}}.
\label{eq:fuse_M_modalities}
\end{align}

{\bf Independence assumptions.} ProbEn is optimal given the independence assumption from~\eqref{eq:ind}. 
Even when such independence assumptions do not hold in practice, the resulting models may still be effective~\cite{dawid1979conditional} (i.e., just as assumptions of Gaussianity can still be useful even if strictly untrue~\cite{kittler1998combining,pearl2014probabilistic}).
Interestingly, many fusion methods including NMS and averaging make the same underlying assumption, 
as discussed in~\cite{kittler1998combining}. In fact, \cite{kittler1998combining} points out that Average Fusion (which averages class posteriors) makes an even stronger assumption: posteriors do not deviate dramatically from class priors. This is likely  not true, as corroborated by the poor performance of averaging in our experiments (despite its apparent widespread use~\cite{xu2014evidential,liu2016multispectral,li2018multispectral}).

{\bf Relationship to prior work.} To compare to prior fusion approaches that tend to operate on logit scores, we rewrite the single-modal softmax posterior for class-$k$ given modality $i$ in terms of single-modal logit score $s_i[k]$.
For notational simplicity, we suppress its dependence on the underlying input modality $x_i$:
$p(\text{$y$$=$$k$}|x_i) = \frac{\exp(s_i[k])}{\sum_j \exp({s_i[j]})}  \ \propto  \ \exp(s_i[k]) \label{logit}$, where we exploit the fact that the partition function in the denominator is not a function of the class label $k$. We now plug the above into Eq.~\eqref{eq:fuse_M_modalities}:
\vspace{-2mm}
\begin{align}\small
    p(\text{$y$$=$$k$}| \{x_i\}_{i=1}^{M}) &\propto \frac{\Pi_{i=1}^{M} p(\text{$y$$=$$k$}|x_i)}{p(\text{$y$$=$$k$})^{M-1}} 
    \propto \frac{\exp(\sum_{i=1}^M s_i[k])}{p(\text{$y$$=$$k$})^{M-1}}
    \label{eq:sum_logits_fuse}
\end{align}
ProbEn is thus equivalent to \emph{summing logits}, dividing by the class prior and normalizing via a softmax. 
Our derivation \eqref{eq:sum_logits_fuse} reveals that summing logits without the division may over-count class priors, where the over-counting grows with the number of modalities $M$. 
The supplement shows that dividing by class posteriors $p(y)$ marginally helps. In practice, we empirically find that assuming uniform priors works surprisingly well, even on imbalanced datasets. This is the default for our experiments, unless otherwise noted.

\begin{figure}[t]
\vspace{-3.2mm}
{\scriptsize \ \hspace{6mm} {\bf (a)} RGB \hspace{14mm} {\bf (b)} thermal \hspace{10mm} {\bf (c)} average fusion \hspace{8mm} {\bf (d)} ProbEn} \hspace{4mm} \ \\
\centering 
\vspace{-0mm}
\includegraphics[width=0.23\linewidth, trim= 6cm 3.8cm 3cm 5cm, clip]{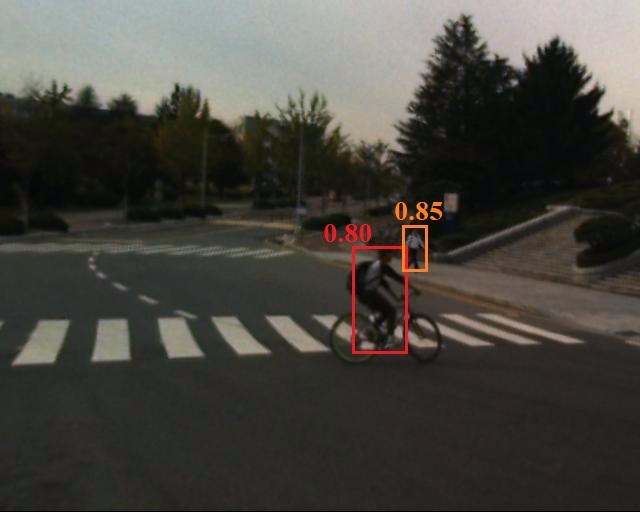}
\includegraphics[width=0.23\linewidth, trim= 6cm 3.8cm 3cm 5cm, clip]{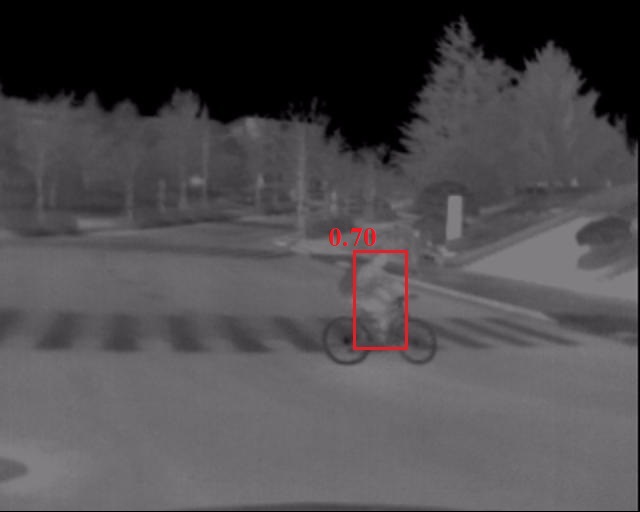}
\includegraphics[width=0.23\linewidth, trim= 6cm 3.8cm 3cm 5cm, clip]{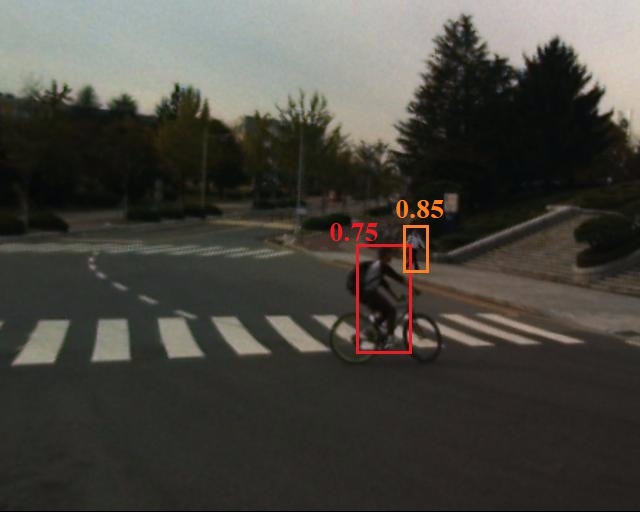}
\includegraphics[width=0.23\linewidth, trim= 6cm 3.8cm 3cm 5cm, clip]{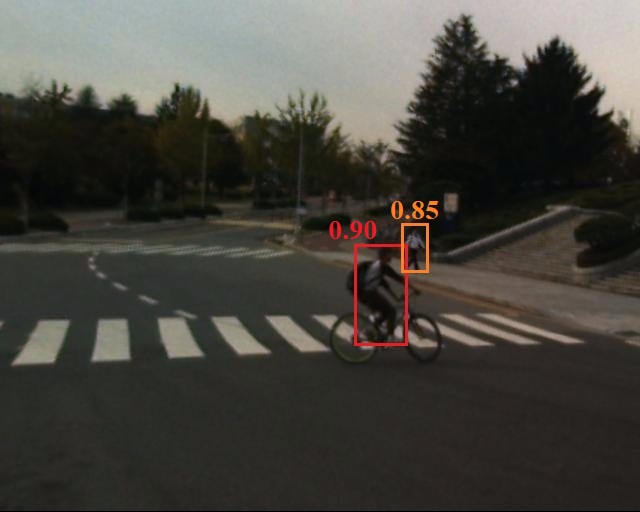}
\vspace{-4mm}
\label{fig:missing_modality}
\caption{\small {\bf Missing modalities.} The {\color{orange} orange-person} {\bf (a)} fails to trigger a thermal detection ({\bf b}), resulting in a single-modal RGB detection (0.85 confidence). To generate an output set of detections (for downstream metrics such as average precision), this detection must be compared to the fused multimodal detection of the {\color{purered}red-person} (RGB: 0.80, thermal: 0.70). {\bf (c)} averaging confidences for the {\color{purered}red-person} lowers their score (0.75) below the {\color{orange} orange-person}, which is unintuitive because additional detections should boost confidence. {\bf (d)} ProbEn increases the {\color{purered}red-person} fused score to 0.90, allowing for proper comparisons to single-modal detections.
}
\vspace{-4mm}
\label{fig:illustration-missing-modality}
\end{figure}

{\bf Missing modalities.} 
Importantly, summing and averaging behave profoundly differently when fusing across ``missing" modalities (Fig.~\ref{fig:illustration-missing-modality}). Intuitively, different single-modal detectors often do not fire on the same object. This means that to output a final set of detections above a confidence threshold (e.g., necessary for computing precision-recall metrics), one will need to compare scores from fused multi-modal detections with single modal detections, as illustrated in Fig.~\ref{fig:illustration-missing-modality}. 
ProbEn elegantly deals with missing modalities because \emph{probabilistically-normalized} multi-modal posteriors $p(y|x_1,x_2)$ can be directly compared with single-modal posteriors $p(y|x_1)$.

{\bf Bounding Box Fusion}.
Thus far, we have focused on fusion of class posteriors. We now extend ProbEn to probabilistically fuse bounding box (bbox) coordinates of overlapping detections. We repurpose the derivation 
from~\eqref{eq:fuse} for a continuous bbox label rather than a discrete one. Specifically, we write $\z$ for the continuous random variable defining the bounding box (parameterized by its centroid, width, and height) associated with a given detection. We assume single-modal detections provide a posterior $p(\z|x_i)$ that takes the form of a Gaussian with a single variance $\sigma_i^2$, i.e., $p(\z|x_i)={\cal N}(\muu_i, \sigma_i^2\I)$
where $\muu_i$ are box coordinates predicted from modality $i$. We also assume a uniform prior on $p(\z)$, implying bbox coordinates can lie anywhere in the image plane. Doing so, we can write  
\vspace{-5mm}
\begin{align}\small
 \ p(\z | x_1, x_2) \propto & \ p(\z | x_1) p(\z|x_2) 
\propto   \ \exp\big(\frac{\Vert \z-\muu_1 \Vert^2}{-2\sigma_1^2} \big) \exp\big(\frac{\Vert \z-\muu_2 \Vert^2}{-2\sigma_2^2} \big) \\
\propto & \ \exp\big(\frac{||\z-\muu||^2}{-2(\frac{1}{\sigma_1^2} + \frac{1}{\sigma_2^2})}), \quad \text{where} \ \  \muu=\frac{\frac{\muu_1}{\sigma_1^2}+\frac{\muu_2}{\sigma_2^2}}{\frac{1}{\sigma_1^2}+\frac{1}{\sigma_2^2}}
\label{eq:bbox-fusion}
\vspace{-4mm}
\end{align}
We refer the reader to the supplement for a detailed derivation.
Eq.~\eqref{eq:bbox-fusion} suggests a simple way to probabilistically fuse box coordinates: compute a weighted average of box coordinates, where weights are given by the inverse covariance. We explore three methods for setting $\sigma^2_i$. The first method ``avg'' fixes $\sigma_i^2$=$1$, amounting to simply averaging bounding box coodinates. 
The second ``s-avg'' approximates $\sigma_i^2 \approx \frac{1}{p(y=k|x_i)}$, implying that more confident detections should have a higher weight when fusing box coordinates. This performs marginally better than simply averaging. 
The third ``v-avg'' train the detector to predict regression \emph{variance}/uncertainty using the Gaussian negative log likelihood (GNLL) loss~\cite{nix1994estimating} alongside the box regression loss. Interestingly,  incorporating GNLL not only produces better variance/uncertainty estimate helpful for fusion but also improves detection performance of the trained detectors (details in supplement).

\section{Experiments}
\begin{figure}[t]
\centering
\begin{minipage}[r]{0.37\textwidth}
\centering
\includegraphics[width=0.7\linewidth, height=1.8cm, trim= 0cm 0cm 8cm 0cm, clip]{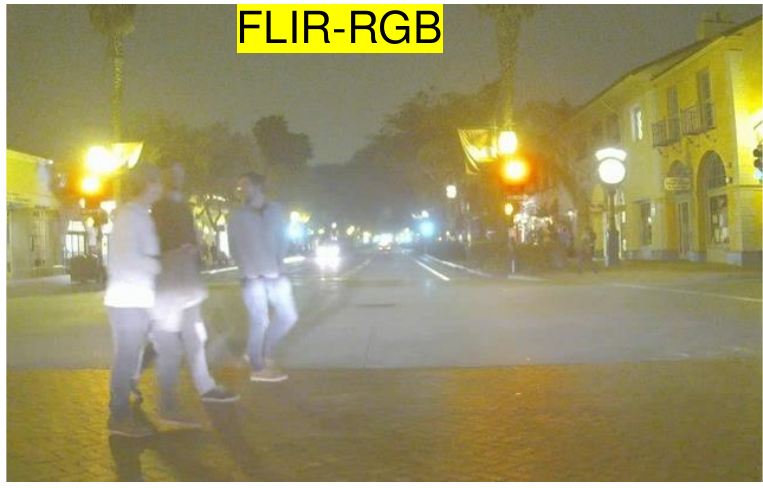} \\
\includegraphics[width=0.7\linewidth, height=1.8cm, trim= 8cm 0cm 0cm 0cm, clip]{figs/FLIR-RGBT.pdf}
\end{minipage} \hfill
\begin{minipage}[c]{0.62\textwidth}
\vspace{-1mm}
\caption{\small 
RGB and thermal images are unaligned both spatially and temporally in FLIR~\cite{FLIR}, which annotates only thermal images.
As a result, prior methods relies on thermal and drop the RGB modality.
We find mid-fusion, taking both RGB and thermal as input, notably improves detection accuracy. When late-fusing detections computed by the mid-fusion and thermal-only detectors, our ProbEn yields much better performance (Table~\ref{tab:ablation_FLIR} and \ref{tab:SOTA_FLIR}).
}
\label{fig:FLIR-annotation}
\end{minipage}
\vspace{-5mm}
\end{figure}

We validate different fusion methods on two datasets: KAIST~\cite{hwang2015multispectral} which is released under the Simplified BSD License, and FLIR~\cite{FLIR} (Fig.~\ref{fig:FLIR-annotation}), which allows for non-commercial educational and research purposes.
Because the two datasets contain personally identifiable information such as faces and license plates,
we assure that we (1) use them only for research, and (2) will release our code and models to the public without redistributing the data.
We first describe implementation details and then report the experimental results on each dataset (alongside their evaluation metrics) in separate subsections.

\subsection{Implementation}
We conduct experiments with PyTorch~\cite{paszke2017automatic} on a single GPU (Nvidia GTX 2080).
We train our detectors (based on Faster-RCNN) with Detectron2~\cite{wu2019detectron2},
using SGD and learning rate 5e-3. 
For data augmentation, we adopt random flipping and resizing.
We pre-train our detector on COCO dataset~\cite{lin2014microsoft}. As COCO has only RGB images, fine-tuning the pre-trained detector on thermal inputs needs careful pre-processing of thermal images (detailed below).

{\bf Pre-processing}. All RGB and thermal images have intensity in [0, 255]. In training an RGB-based detector, RGB input images are commonly processed using the mean subtraction~\cite{wu2019detectron2} where the mean values are computed over all the training images.
Similarly, we calculate the mean value (135.438) in the thermal training data. We find using a precise mean subtraction to process thermal images yields better performance when fine-tuning the pre-trained detector.

{\bf Stage-wise Training}.
We fine-tune the pre-trained detector to train single-modal detectors and the early-fusion detectors.
To train a mid-fusion detector, we truncate the {\em already-trained} single-modal detectors, concatenate features add a new detection head and train the whole model (Fig.~\ref{fig:flowchart}a).
The late-fusion methods fuse detections from (single-modal) detectors. Note that all the late-fusion methods are {\em non-learned}. 
We also experimented with learning-based late-fusion methods (e.g., learning to fuse logits) but find them to be only marginally better than ProbEn (9.08 vs. 9.16 in LAMR using argmax box fusion). Therefore, we focus on the non-learned late fusion methods in the main paper and study learning-based ones in the supplement.

{\bf Post-processing}.
When ensembling two detectors, we find it crucial to calibrate scores particularly when we we fuse detections from our in-house models and off-the-shelf models released by others. 
We adopt the simple temperature scaling for score calibration~\cite{guo2017calibration}. Please refer to the supplement for details.

\begin{figure}[t]
\centering
\begin{minipage}[r]{0.47\textwidth}
\centering
\includegraphics[width=0.48\linewidth, trim= 0cm 0cm 11.5cm 0cm, clip]{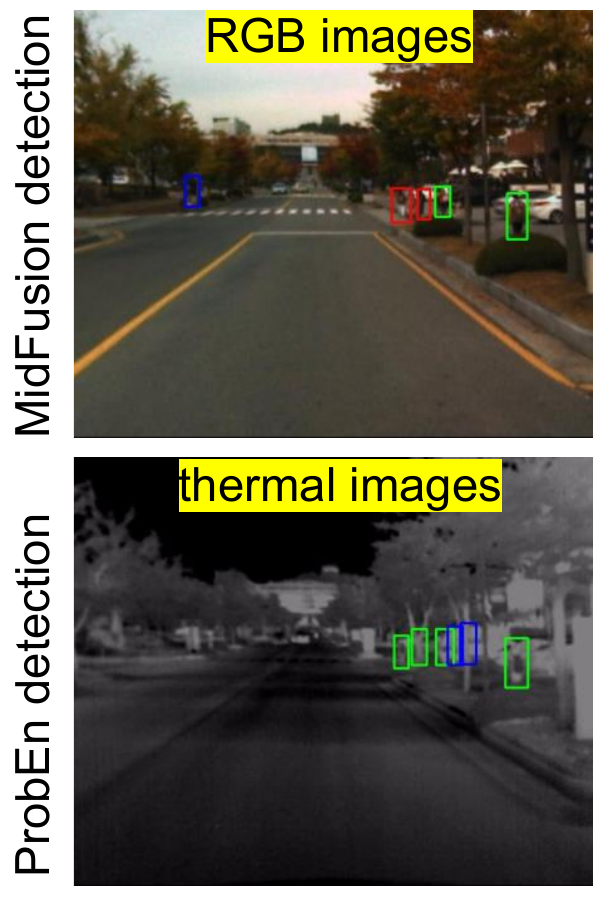}
\includegraphics[width=0.4485\linewidth, trim= 11.9cm 0cm 0cm 0cm, clip]{figs/KAIST-RGBT-results.pdf}
\end{minipage} \hfill
\begin{minipage}[c]{0.50\textwidth}
\vspace{-3mm}
\caption{\small 
Detections overlaid on two KAIST testing examples in columns.
{\bf Top}: detections by our mid-fusion model.
{\bf Bottom}: detections by our ProbEn by fusing detections of thermal-only and mid-fusion models. 
{\color{darkgreen} Green}, {\color{red} red} and {\color{darkblue} blue} boxes stand for {\color{darkgreen} true positives}, {\color{red} false negative} (miss-detection) and {\color{darkblue} false positives}. 
Visually, ProbEn performs much better than the mid-fusion model, which is already comparable to the prior work as shown in Table~\ref{tab:ablation_KAIST} and~\ref{tab:SOTA_KAIST}.
}
\label{fig:KAIST-RGBT-results}
\end{minipage}
\vspace{-4mm}
\end{figure}

\subsection{Multimodal Pedestrian Detection on KAIST}

{\bf Dataset}.
The KAIST dataset is a popular multimodal benchmark for pedestrian detection~\cite{hwang2015multispectral}.
In KAIST, RGB and thermal images are aligned with a beam-splitter, and have resolutions of 640x480 and 320x256, respectively.
We resize thermal images to 640x480 during training. KAIST also provides day/night tags for breakdown analysis.
The original KAIST dataset contains 95,328 RGB-thermal image pairs, which are split into a training set (50,172) and a testing set (45,156). 
Because the original KAIST dataset contains noisy annotations, the literature introduces cleaned version of the train/test sets: a sanitized train-set (7,601 examples)~\cite{li2018multispectral} and a cleaned test-set (2,252 examples)~\cite{liu2018improved}.
We also follow the literature~\cite{hwang2015multispectral} to evaluate under the ``reasonable setting'' for evaluation by ignoring annotated persons that are occluded (tagged by KAIST) or too small ($<$55 pixels). 
We follow this literature for fair comparison with recent methods.

{\bf Metric}.
We measure detection performance with the Log-Average Miss Rate (LAMR), which is a standard metric in pedestrian detection~\cite{dollar2011pedestrian} and KAIST~\cite{hwang2015multispectral}. 
LAMR is computed by averaging the miss rate (false negative rate) at nine false positives per image (FPPI) rates evenly spaced in log-space from the range 10$^{-2}$ to 10$^0$~\cite{hwang2015multispectral}.
It does not evaluate the detections that match to ignored ground-truth~\cite{dollar2011pedestrian,hwang2015multispectral}.
A true positive is a detection that matches a ground-truth object with IoU$>$0.5~\cite{hwang2015multispectral};
false positives are detections that do not match any ground-truth;
false negatives are miss-detections.

\subsubsection{Ablation Study on KAIST}

{
\setlength{\tabcolsep}{0.25em} 
\begin{table}[t]
\small
\vspace{-1mm}
\scalebox{0.83}{
\begin{tabular}{l l ccc}
\toprule
\multicolumn{2}{l}{\cellcolor{lightlightgrey}baselines}  &  \cellcolor{lightlightgrey}{\em Day} & \cellcolor{lightlightgrey}{\em Night} & \cellcolor{lightlightgrey}{\em All} \\
\midrule
\multicolumn{2}{l}{RGB}                     & 14.56 & 27.42 & 18.67 \\
\multicolumn{2}{l}{Thermal}                 & 24.59 & 7.76 & 18.99 \\ 
\multicolumn{2}{l}{EarlyFusion}             & 26.30 & 6.61 & 19.36 \\
\multicolumn{2}{l}{MidFusion}              & 17.55 & 9.30 & 14.48 \\
\multicolumn{2}{l}{Pooling}                 & 37.92 & 22.61 & 32.68 \\
\midrule
\cellcolor{lightlightgrey}{\em score-fusion} & \cellcolor{lightlightgrey}{\em box-fusion} &  \cellcolor{lightlightgrey}{\em Day} & \cellcolor{lightlightgrey}{\em Night} & \cellcolor{lightlightgrey}{\em All} \\
\midrule
max                 &  argmax& 13.25 & 6.42 & 10.78 \\
max                 &  avg   & 13.25 & 6.65 & 10.89 \\
max                 &  s-avg & 13.35 & 6.65  & 10.96 \\
max                 &  v-avg & 13.19 & 6.65  & 10.79 \\
avg                 &  argmax& 21.68 & 15.16 & 19.53 \\
avg                 &  avg   & 21.59 & 15.46 & 19.47 \\
avg                 &  s-avg & 21.67 & 15.46 & 19.55 \\
avg                 &  v-avg & 21.51 & 15.46 & 19.42 \\
ProbEn              &  argmax& 10.21 & 5.45 & 8.62 \\ 
ProbEn              &  avg   & 10.14 & 5.41 & 8.58 \\ 
ProbEn              &  s-avg & 10.27 & 5.41 & 8.67 \\ 
ProbEn              &  v-avg & 9.93 & 5.41 & 8.50 \\
ProbEn$_3$             &   argmax   & 13.67 & 6.31 & 11.00 \\ 
ProbEn$_3$             &   avg   & 9.07 & 4.89 & 7.68 \\ 
ProbEn$_3$             &  s-avg   & {\bf 9.07} & {\bf 4.89} & 7.68 \\ 
ProbEn$_3$             &  v-avg   & {\bf 9.07} & {\bf 4.89} & {\bf 7.66} \\ 
\bottomrule
\end{tabular}
}
\begin{minipage}[r]{0.55\textwidth}
\vspace{2mm}
\centering
\caption{\small 
{\bf Ablation study on KAIST} (LAMR$\downarrow$ in \%). 
The upper panel shows that (1) RGB-only and Thermal-only detectors perform notably better than each other on \emph{Day} and \emph{Night} respectively, and (2) MidFusion strikes a balance and performs better overall.
In the lower panel, we focus on the very-late fusion of RGB and Thermal. We ablate methods for \emph{score fusion} (max as in NMS, avg and ProbEn), and \emph{box fusion} (argmax as in NMS, ProbEn that uses avg, s-avg or v-avg).
Somewhat surprisingly, ``max + argmax'', or NMS, performs quite well on both \emph{Day} and \emph{Night}; average score fusion performs poorly because it double counts class prior.
As for box fusion, using the learned variance / uncertainty by v-avg performs better than the heuristic methods (avg and s-avg).
Our ProbEn performs significantly better and ProbEn$_3$ is the best by fusing three models: RGB, Thermal, and MidFusion.
}
\label{tab:ablation_KAIST}
\end{minipage} 
\vspace{-6mm}
\end{table}
}

Table~\ref{tab:ablation_KAIST} shows ablation studies on KAIST.
Single modal detectors tend to work well in different environments, with RGB detectors working on well-lit day images while Thermal working well on nighttime images. EarlyFusion reduces the miss rate by a modest amount, while MidFusion is more effective. Naive strategies for late fusion (such as pooling together detections from different modalities) are quite poor because they generate many repeated detections on the same object, which are counted as false positives. Interestingly, simple NMS that has max score fusion and argmax box fusion, is quite effective at removing overlapping detections from different modalities, already outperforming Early and MidFusion. Instead of suppressing the weaker modality,  one might average the scores of overlapping detections but this is quite ineffective because it always decreases the score from NMS. 
Intuitively, one should increase the score when different modalities agree on a detection. ProbEn accomplishes this by probabilistic integration of information from the RGB and Thermal single-modal detectors. 
Moreover, it can be further improved by probabilisitcally fusing coordinates of overlapping boxes.
Lastly, ProbEn$_3$ that ensembles three models (RGB, thermal and MidFusion), performs the best.

{\bf Qualitative Results} are displayed in Fig.~\ref{fig:KAIST-RGBT-results}.
Visually, ProbEn detects all persons, while the MidFusion model has multiple false negatives /  miss-detections.

\subsubsection{Quantitative Comparison on KAIST}

{\bf Compared Methods}.
Among many prior methods, we particularly compare against four recent ones:
AR-CNN~\cite{zhang2019weakly}, MBNet~\cite{zhou2020improving}, MLPD~\cite{kim2021mlpd}, and GAFF~\cite{zhang2021guided}.
AR-CNN focuses on weakly-unaligned RGB-thermal pairs and explores multiple heuristic methods for fusing features, scores and boxes.
MBNet addresses modality imbalance w.r.t illumination and features to improve detection; both MLPD and GAFF are mid-fusion methods that design sophisticated network architectures; MLPD adopts aggressive data augmentation techniques and GAFF extensively exploits attentitive modules to fuse multimodal features. Table~\ref{tab:SOTA_KAIST} lists more methods.

{\bf Results}.
Table~\ref{tab:SOTA_KAIST} compares ProbEn against the prior work. ProbEn+ that ensembles three models trained in-house (RGB, Thermal, and MidFusion) achieves competitive performance (7.95 LAMR) against the prior art.
When replacing our MidFusion detector with off-the-shelf mid-fusion detectors~\cite{kim2021mlpd,zhang2021guided},
ProbEn++ significantly outperforms all the existing methods, boosting the performance from the prior art 6.48 to 5.14!
This clearly shows that ProbEn works quite well when the conditional independence assumption does not hold, i.e., fusing outputs from other fusion methods (both off-the-shelf and trained in-house).
As ProbEn performs better than past work as a non-learned solution, we argue that it should serve as a new baseline for future research on multimodal detection.

{
\setlength{\tabcolsep}{0.10em} 
\begin{table}[t]
\small
\scalebox{0.9}{
\begin{tabular}{l|c|c|c}
\hline
{\em Method} & {\em Day} & {\em Night} & {\em All} \\
\hline
HalfwayFusion~\cite{liu2016multispectral} & 36.84 & 35.49 &36.99 \\ 
RPN+BDT~\cite{konig2017fully} & 30.51 & 27.62 & 29.83 \\ 
TC-DET~\cite{bertinitask} & 34.81 & 10.31 & 27.11 \\
IATDNN~\cite{guan2019fusion} & 27.29 & 24.41 & 26.37 \\ 
IAF R-CNN~\cite{li2019illumination} & 21.85 & 18.96 & 20.95 \\
SyNet~\cite{albaba2021synet}    & 22.64 & 15.80  & 20.19 \\
CIAN~\cite{zhang2019cross} & 14.77 & 11.13 & 14.12 \\
MSDS-RCNN~\cite{li2018multispectral} & 12.22 & 7.82 & 10.89 \\
AR-CNN~\cite{zhang2019weakly} & 9.94 & 8.38 & 9.34 \\
MBNet~\cite{zhou2020improving} & 8.28 & 7.86 & 8.13 \\
MLPD~\cite{kim2021mlpd} & 7.95 & 6.95 & 7.58 \\
GAFF~\cite{zhang2021guided} & 8.35 & 3.46 & 6.48 \\
\hline
\cellcolor{lightlightgrey}{MaxFusion (NMS)}               &
\cellcolor{lightlightgrey}{13.25} & 
\cellcolor{lightlightgrey}{6.42} & 
\cellcolor{lightlightgrey}{10.78} \\
\cellcolor{lightlightgrey}{ProbEn}               &
\cellcolor{lightlightgrey}{9.93} &    
\cellcolor{lightlightgrey}{5.41} & 
\cellcolor{lightlightgrey}{8.50} \\
\cellcolor{lightlightgrey}{ProbEn$_3$}  & \cellcolor{lightlightgrey}{9.07} & \cellcolor{lightlightgrey}{4.89} & \cellcolor{lightlightgrey}{7.66} 
\\
\cellcolor{lightlightgrey}{ProbEn$_3$ w/ MLPD}  & \cellcolor{lightlightgrey}{7.81} & \cellcolor{lightlightgrey}{5.02} & \cellcolor{lightlightgrey}{6.76} 
\\
\cellcolor{lightlightgrey}{ProbEn$_3$ w/ GAFF}  & \cellcolor{lightlightgrey}\textbf{6.04} & \cellcolor{lightlightgrey}\textbf{3.59} & \cellcolor{lightlightgrey}\textbf{5.14} \\
\hline
\end{tabular}
}
\begin{minipage}[r]{0.58\textwidth}
\small
\centering
\vspace{3mm}
\caption{\small 
{\bf Benchmarking on KAIST}  measured by \% LAMR$\downarrow$.
We report numbers from the respective papers.
Results are comparable to  Table~\ref{tab:ablation_KAIST}. 
{\em Simple probabilistic ensembling of independently-trained detectors (ProbEn) outperforms $\frac{9}{12}$ methods on the leaderboard. Infact, even NMS (MaxFusion) outperforms $\frac{8}{12}$ methods, indicating the under-appreciated effectiveness of detector-ensembling as a multimodal fusion technique.} Performance further increases when adding a MidFusion detector to the probabilistic ensemble ($\text{ProbEn}_3$).
Replacing our in-house MidFusion with off-the-shelf mid-fusion detectors MLPD~\cite{kim2021mlpd} and GAFF~\cite{zhang2021guided} significantly boosts the state-of-art from 6.48 to 5.14!
This shows ProbEn remains effective even when fusing models for which conditional independence does not hold. 
}
\label{tab:SOTA_KAIST}
\end{minipage}
\vspace{-7mm}
\end{table}
}

\subsection{Multimodal Object Detection on FLIR}

{\bf Dataset}.
The FLIR dataset~\cite{FLIR} consists of RGB images (captured by a FLIR BlackFly RGB camera with 1280x1024 resolution) and thermal images (acquired by a FLIR Tau2 thermal camera 640x512 resolution). 
We resize all images to resolution 640x512.
FLIR has $10,228$ \emph{unaligned} RGB-thermal image pairs and annotates only for thermal  (Fig.~\ref{fig:FLIR-annotation}).
Image pairs are split into train-set ($8,862$ images) and a validation set ($1,366$ images).
FLIR evaluates on three classes which have imbalanced examples~\cite{cao2019every,kieubottom,zhang2020multispectral,munir2020thermal,devaguptapu2019borrow}:
$28,151$ persons, $46,692$ cars, and $4,457$ bicycles. 
Following~\cite{zhang2020multispectral}, we remove 108 thermal images in the val-set that do not have the RGB counterparts.
For breakdown analysis w.r.t day/night scenes, we manually tag the validation images with ``day'' (768) and ``night'' (490). We will release our annotations to the public.

{\bf Misaligned modalities}. 
Because FLIR's RGB and thermal images are heavily unaligned, it labels only thermal images and does not have RGB annotations.
We can still train Early and MidFusion models using multimodal inputs and the thermal annotations.
These detectors might learn to internally align the unaligned modalities to predict bounding boxes according to the thermal annotations.
Because we do not have an RGB-only detector, our ProbEn ensembles EarlyFusion, MidFusion, and thermal-only detectors.

{\bf Metric}. 
We measure performance using Average Precision (AP)~\cite{everingham2015pascal,russakovsky2015imagenet}. 
Precision is computed over testing images within a single class, with true positives that overlap ground-truth bounding boxes (e.g., IoU$>$0.5).
Computing the average precision (AP) across all classes measures the performance in multi-class object detection.
Following~\cite{devaguptapu2019borrow,munir2020thermal,zhang2020multispectral,kieubottom,cao2019every}, we define a true positive as a detection that overlaps a ground-truth with IoU$>$0.5. 
Note that AP used in the the multimodal detection literature is different from mAP~\cite{lin2014microsoft}, which averages over different AP's computed with different IoU thresholds.

{
\setlength{\tabcolsep}{0.43em} 
\begin{table}[t]
\small
\vspace{-0mm}
\scalebox{0.80}{
\begin{tabular}{l l ccc}
\toprule
\multicolumn{2}{l}{\cellcolor{lightlightgrey}baselines}  &  \cellcolor{lightlightgrey}{\em Day} & \cellcolor{lightlightgrey}{\em Night} & \cellcolor{lightlightgrey}{\em All} \\
\midrule
\multicolumn{2}{l}{Thermal}                 & 75.35 & 82.90 & 79.24 \\ 
\multicolumn{2}{l}{EarlyFusion}             & 77.37 & 79.56 & 78.80 \\ 
\multicolumn{2}{l}{MidFusion}              & 79.37 & 81.64 & 80.53 \\
\multicolumn{2}{l}{Pooling}                 & 52.57  & 55.15  & 53.66  \\
\midrule
\cellcolor{lightlightgrey}{\em score-fusion} & \cellcolor{lightlightgrey}{\em box-fusion} &  \cellcolor{lightlightgrey}{\em Day} & \cellcolor{lightlightgrey}{\em Night} & \cellcolor{lightlightgrey}{\em All} \\
\midrule
max                 &  argmax& 81.91 & 84.42 & 83.14 \\
max                 &  avg   & 81.84 & 84.62 & 83.21 \\
max                 &  s-avg & 81.85 & 84.48 & 83.19 \\
max                 &  v-avg & 81.80 & 85.07 & 83.31 \\
avg                 &  argmax& 81.34 & 84.69 & 82.65 \\
avg                 &  avg   & 81.26 & 84.81 & 82.91 \\
avg                 &  s-avg & 81.26 & 84.72 & 82.89 \\
avg                 &  v-avg & 81.26 & 85.39 & 83.03 \\
ProbEn$_3$              &  argmax& 82.19 & 84.73 & 83.27 \\ 
ProbEn$_3$              &  avg   & 82.19 & {84.91} & 83.63 \\ 
ProbEn$_3$              &  s-avg & 82.20 & 84.84 & 83.61 \\ 
ProbEn$_3$              &  v-avg & {\bf 82.21}       & {\bf 85.56} & {\bf 83.76}\\
\bottomrule
\end{tabular}
}
\begin{minipage}[r]{0.5\textwidth}
\centering
\vspace{3mm}
\caption{\small 
{\bf Ablation study on FLIR} day/night scenes (AP$\uparrow$  in percentage with IoU$>$0.5).
Compared to thermal-only detector, incorporating RGB by EarlyFusion and MidFusion notably improves performance.
Late-fusion (lower panel) ensembles three detectors: Thermal, EarlyFusion and MidFusion.
All the explored late-fusion methods lead to better performance than MidFusion. In particular,  ProbEn performs the best.
Moreover, similar to the results on KAIST, 
using predicted uncertainty to fuse boxes (v-avg) performs better than the other two heuristic box fusion methods, avg that naively averages box coordinates and s-avg that uses classification scores to weighted average box coordinates.
}
\label{tab:ablation_FLIR}
\end{minipage}
\vspace{-5mm}
\end{table}
}

\subsubsection{Ablation study on FLIR}

\begin{figure}[t]
\centering
\begin{minipage}[l]{0.47\textwidth}
    \centering
\hspace{4mm} {\small day scene} \hspace{11mm}  {\small night scene} \\
\includegraphics[width=0.53\linewidth, trim= 0cm 0cm 12cm 0cm, clip]{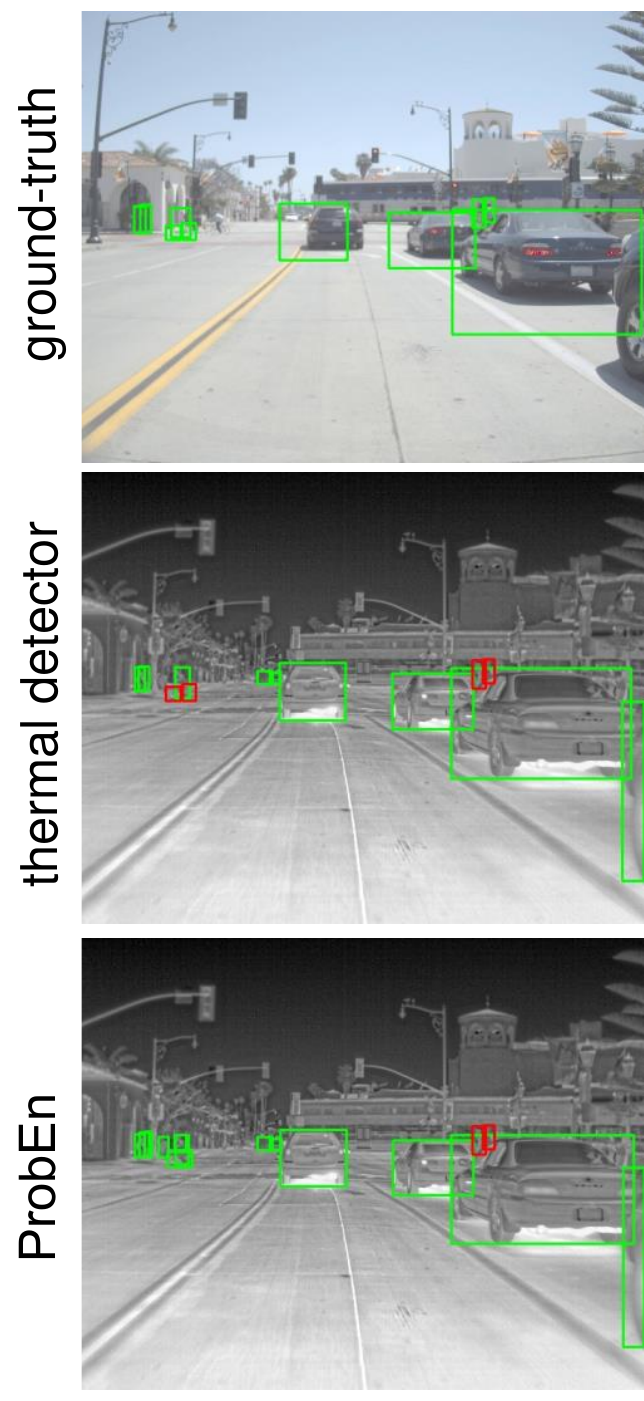}
\includegraphics[width=0.45\linewidth, trim= 13cm 0cm 0cm 0cm, clip]{figs/FLIR-visual-results.pdf}
\end{minipage} \hfill
\begin{minipage}[r]{0.52\textwidth}
\caption{\small 
Detections overlaid on two FLIR testing images (in columns) with RGB (top) and thermal images (middle and bottom). To avoid clutter, we do not mark class labels for the bounding boxes.
Ground-truth annotations are shown on the RGB, emphaszing that RGB and thermal images are strongly unaligned. 
On the thermal images, we compare thermal-only (mid-row) and our ProbEn (bottom-row) models. 
{\color{darkgreen} Green}, {\color{red} red} and {\color{darkblue} blue} boxes stand for {\color{darkgreen} true positives}, {\color{red} false negative} (mis-detected persons) and {\color{darkblue} false positives}. 
In particular, in the second column, the thermal-only model has many false negatives (or miss-detections), which are ``bicycles''. Understandably, thermal cameras will not capture bicycles because they do not emit heat. In contrast, RGB capture bicycle signatures better than thermal. This explains why our fusion performs better on  bicycles.
}
\label{fig:FLIR-RGBT-results}
\end{minipage}
\vspace{-5mm}
\end{figure}

We compare our fusion methods in  Table~\ref{tab:ablation_FLIR}, along with qualitative results in Fig.~\ref{fig:FLIR-RGBT-results}.
We analyze results using our day/night tags.
Compared to the single-modal detector (Thermal), our learning-based early-fusion (EarlyFusion) and mid-fusion (MidFusion) produce better performance.
MidFusion outperforms EarlyFusion, implying that end-to-end learning of fusing features better handles mis-alignment between RGB and thermal images.
By applying late-fusion methods to detections of Thermal, EarlyFusion and MidFusion detectors, we boost detection performance.
Note that typical ensembling methods in the single-modal (RGB) detection literature~\cite{xu2014evidential,liu2016multispectral,li2018multispectral} often use max / average score fusion, and argmax / average box fusion,
which are outperformed by our ProbEn. 
This suggests that ProbEn should be potentially a better ensembling method for object detection.

\subsubsection{Quantitative Comparison on FLIR}
{\bf Compared Methods}.
We compare against prior methods including ThermalDet~\cite{cao2019every}, BU~\cite{kieubottom}, ODSC~\cite{munir2020thermal}, MMTOD~\cite{devaguptapu2019borrow}, CFR~\cite{zhang2020multispectral}, and GAFF~\cite{zhang2021guided}. 
As FLIR does not have aligned RGB-thermal images and only annotates thermal images, many methods exploit domain adaptation that adapts a pre-trained RGB detector to thermal input.
For example, MMTOD~\cite{devaguptapu2019borrow} and ODSC~\cite{munir2020thermal} adopt the image-to-image-translation technique~\cite{zhu2017unpaired,zhang2017multi} to generate RGB from thermal,  hypothesizing that this helps train a better multimodal detector by finetuning a detector that is pre-trained over large-scale RGB images.
BU~\cite{kieubottom} operates such a translation/adaptation on features that generates thermal features to be similar to RGB features.
ThermalDet~\cite{cao2019every} exclusively exploits thermal images and ignores RGB images; it proposes to combine features from multiple layers for the final detection.
GAFF~\cite{zhang2021guided} trains on RGB-thermal image with a sophisticated attention module that fuse single-modal features.
Perhaps because the complexity of the attention module, GAFF is limited to using small network backbones (ResNet18 and VGG16). 
Somewhat surprisingly, to the best of our knowledge, there is no prior work that trained early-fusion or mid-fusion deep networks (Fig.~\ref{fig:flowchart}a) on the heavily unaligned RGB-thermal image pairs (like in FLIR) for multimodal detection. We find directly training them performs much better than prior work (Table~\ref{tab:SOTA_FLIR}).

{\bf Results.} 
Table~\ref{tab:SOTA_FLIR} shows that all our methods outperform the prior art.
Our single-modal detector (trained on thermal images) achieves slightly better performance than ThermalDet~\cite{cao2019every}, which also exclusively trains on thermal images.
This is probably because we use a better pre-trained Faster-RCNN model provided by the excellent Detectron2 toolbox.  
Surprisingly, our simpler EarlyFusion and MidFusion models achieve big boosts over the thermal-only model (Thermal), while MidFusion  performs much better.
This confirms our hypothesis that fusing features better handles mis-alignment of RGB-thermal images than the early-fusion method.
Our ProbEn performs the best, significantly better than all compared methods!
Notably, our fusion methods boost ``bicycle'' detection. We conjecture that bicycles do not emit heat to deliver strong signatures in thermal, but are more visible in RGB; fusing them greatly improves bicycle detection.

{
\setlength{\tabcolsep}{0.32em} 
\begin{table}[t]
\vspace{-2mm}
\scalebox{0.8}{
\begin{tabular}{l|c|c|c|c}
\hline
{\em Method} & {\em Bicycle} & {\em Person} & {\em Car} & {\em All} \\
\hline
MMTOD-CG~\cite{devaguptapu2019borrow} & 50.26 & 63.31 & 70.63 & 61.40 \\ 
MMTOD-UNIT~\cite{devaguptapu2019borrow} & 49.43 & 64.47 & 70.72 & 61.54 \\ 
ODSC~\cite{munir2020thermal} & 55.53 & 71.01 &82.33 & 69.62 \\
CFR3~\cite{zhang2020multispectral} & 55.77 & 74.49 & 84.91 & 72.39 \\ 
BU(AT,T)~\cite{kieubottom} & 56.10 & 76.10 & 87.00 & 73.10 \\ 
BU(LT,T)~\cite{kieubottom} & 57.40 & 75.60 & 86.50 & 73.20 \\ 
GAFF~\cite{zhang2021guided} & --- & --- & --- & 72.90 \\ 
ThermalDet~\cite{cao2019every} & 60.04 & 78.24 & 85.52 &74.60 \\ 
\hline
\cellcolor{lightlightgrey}Thermal & \cellcolor{lightlightgrey}62.63 & \cellcolor{lightlightgrey}84.04 & \cellcolor{lightlightgrey}87.11 & \cellcolor{lightlightgrey}79.24\\
\cellcolor{lightlightgrey}EarlyFusion  & \cellcolor{lightlightgrey}63.43 & \cellcolor{lightlightgrey}85.27 & \cellcolor{lightlightgrey}87.69 & \cellcolor{lightlightgrey}78.80\\
\cellcolor{lightlightgrey}MidFusion & \cellcolor{lightlightgrey}69.80 & \cellcolor{lightlightgrey}84.16 & \cellcolor{lightlightgrey}87.63 & \cellcolor{lightlightgrey}80.53\\
\cellcolor{lightlightgrey}ProbEn$_3$  & \cellcolor{lightlightgrey}\textbf{73.49}  & \cellcolor{lightlightgrey}\textbf{87.65}  & \cellcolor{lightlightgrey}\textbf{90.14}  & \cellcolor{lightlightgrey}\textbf{83.76} \\
\hline
\end{tabular}
}
\begin{minipage}[r]{0.47\textwidth}
\vspace{3mm}
\centering
\small
\caption{\small
{\bf Benchmarking on FLIR} measured by AP$\uparrow$ in percentage with IoU$>$0.5 with breakdown on the three categories. 
Perhaps surprisingly, end-to-end training on thermal already outperforms all the prior methods, presumably because of using a better pre-trained model (Faster-RCNN).
Importantly, our ProbEn increases AP from prior art 74.6\% to 84.4\%!
These results are comparable to Table~\ref{tab:ablation_FLIR}.
}
\label{tab:SOTA_FLIR}
\end{minipage}
\vspace{-9mm}
\end{table}
}

\section{Discussion and Conclusions}

We explore different fusion strategies for multimodal detection under both aligned and unaligned RGB-thermal images. 
We show that non-learned probabilistic fusion, ProbEn, significantly outperforms prior  approaches. Key reasons for its strong performance are that (1) it can take advantage of highly-tuned single-modal detectors trained on large-scale single-modal datasets, and (2) it can deal with missing detections from particular modalities, a common occurrence when fusing together detections. 
One by-product of our diagnostic analysis is the remarkable performance of NMS as a fusion technique, precisely because it exploits the same key insights.
Our ProbEn yields $>${\bf 13\%} relative improvement over prior work, both on aligned and unaligned multimodal benchmarks.

{\small
\
\\
\noindent{\bf Acknowledgement}.
This work was supported by the CMU Argo AI Center for Autonomous Vehicle Research. Authors acknowledge valuable discussions with Jessica Lee, Peiyun Hu, Jianren Wang, David Held, Kangle Deng, and Michel Laverne.
}

\bibliographystyle{splncs04}
\bibliography{egbib}

\newpage
\section*{}
\begin{center}
{\bf \large Appendix}
\end{center}
The appendix provides additional studies about the proposed probabilistic ensembling technique (ProbEn).
Below is a sketch of document and we refer the reader to each of these sections for details.
\begin{itemize}
    \item Section~\ref{sec:discussion-late-fusion}: Analysis of ProbEn and comparisons to other late-fusion methods.
    \item Section~\ref{sec:score-calibration}: Score calibration for ProbEn
    \item Section~\ref{sec:weighted-fusion}: Further study of weight score fusion
    \item Section~\ref{sec:class-prior}: Further study of class prior in ProbEn
    \item Section~\ref{sec:box-fusion}: A detailed derivation of probabilistic box fusion
    \item Section~\ref{sec:quantitative-results}: A study of fusing more and better models
    \item Section~\ref{sec:visual-results}: Qualitative results and video demo
\end{itemize}

\section{Probabilistic Fusion for Logits}
\label{sec:discussion-late-fusion}

We  compare ProbEn to additional late fusion approaches in the literature that extends beyond detection. Because classic fusion approaches \cite{simonyan2014two,yue2015beyond,Feichtenhofer_2016_CVPR} often operate on logit scores that are input into a softmax (rather than operating on the output of a softmax), we re-examine ProbEn in terms of logit scores.

Let us rewrite the single-modal softmax posterior for class $k$ given modality $i$ in terms of single-modal logit scores $s_i[k]$. For notational simplicity, we suppress its dependence on the underlying input modality $x_i$:
\begin{align}
    p(y=k|x_i) = \frac{e^{s_i[k]}}{\sum_j e^{s_i[j]}} 
    \propto e^{s_i[k]} 
    \label{logit}
\end{align}
where we exploit the fact that the partition function in the denominator is not a function of the class label $k$. We now plug the above into Eq.~\ref{eq:sum_logits_fuse}:
\begin{align}
    p(y=k|x_1,x_2) &\propto \frac{p(y=k|x_1)p(y=k|x_2)}{p(y=k)} \propto \frac{e^{s_1[k] + s_2[k]}}{p(y=k)} \label{eq:sumfuse}
\end{align}

\begin{figure*}[t]
\centering
\includegraphics[width=1\linewidth]{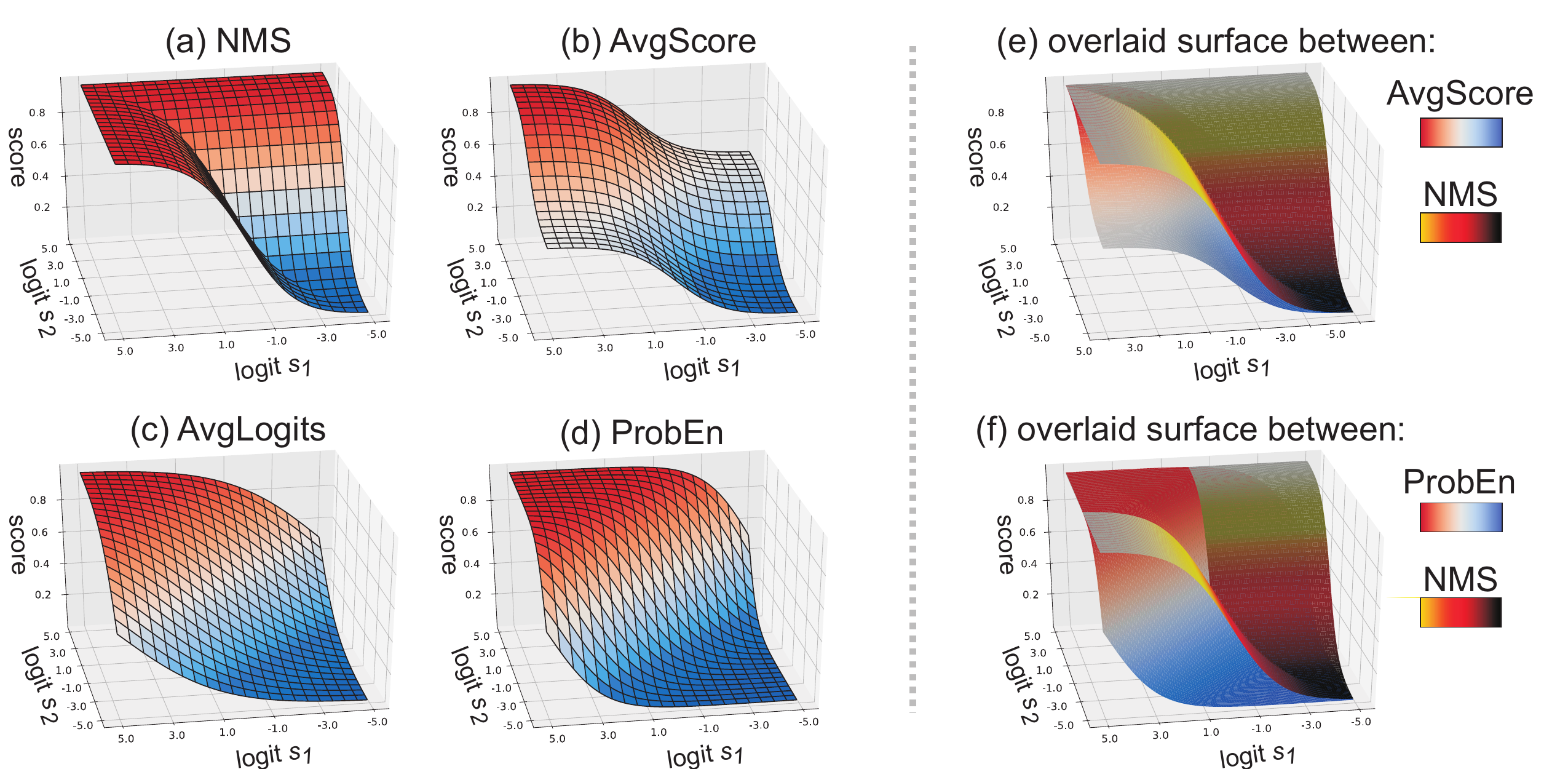}
\vspace{-7mm}
\caption{\small
{\bf Fusing logits from two single-modal, single-class detectors}.
Given a single class detector $k \in \{0,1\}$, the single-modal class posterior for modality $i$ depends on the relative logit $s_i = s_i[1] - s_i[0]$. We visualize the probability surface obtained by different fusion strategies that operate on logit scores $s_1$ and $s_2$ (associated with two overlapping detections).
We first point out that simply returning the maximum score, corresponding to non-maximal suppression (NMS), is a surprisingly effective late fusion strategy that already outperforms much prior work (see Table 1 from main paper and Table~\ref{tab:discussion-late-fusion} in appendix). AvgLogits (c) and ProbEn (d) have similar score landscapes, but differ in a scaling parameter.
Our empirical results show that this scaling parameter has a {\em large} effect in multimodal detection, because one needs to compare multi-modal detections with single-modal detections with ``missing data modalities". 
By overlaying the score landscapes of NMS and AvgScore (e), one can see that AvgScore is always less than NMS.
Similarly, by overlaying the score landscapes of ProbEn and NMS (f), we find that ProbEn returns (1) a higher probability than NMS when both modalities have large logits (e.g., $s_1$=4 and $s_2$=4) but (2) a lower probability than NMS when logit scores disagree (e.g., $s_1=3$ and $s_2=-3$, corresponding to $p(y=1|x_1)=0.95$ and $p(y=1|x_2)=0.05$). In the latter case, NMS outputs an over-confident score 0.95; ProbEn decreases the score, which helps reduce false positives as illustrated in Fig.~\ref{tab:handle-FPs}.
}
\vspace{-3mm}
\label{fig:landspaces}
\end{figure*}

If we assume a uniform prior over classes, Bayesian posteriors are proportional to $e^{s[k]}$ where $s[k] = s_1[k] + s_2[k]$ are the summed per-modality logits. Hence, {\em ProbEn corresponds to adding logits from each modality.} This suggests another practical implementation of ProbEn that may improve numerical stability: given single-modal detections with cached logit scores, sum logit scores on overlapping detections before pushing them through a softmax.

{
\setlength{\tabcolsep}{1.05em} 
\begin{table}[t]
\small
\centering
\caption{ \small
{\bf Additional late fusion baselines} measured by LAMR$\downarrow$ on KAIST reasonable-test. 
Numbers are identical to Table 1 from the main paper with an additional row for logit averaging (AvgLogits), which outperforms class-posterior averaging (AvgScore). However, both methods underperform a simple NMS (MaxFusion). Eq.\eqref{eq:sumfuse} derives that ProbEn is equivalent to {\em summing} logits instead of averaging. Intuitively, summing allows fusion to become more confident as more modalities agree, while averaging does not. Even more importantly, this small modification allows one to properly compare detections with missing modalities, which is frequently needed in NMS whenever all modalities fail to fire on a given object. Finally, we also explore a learned late fusion baseline that learns to combine logits with logistic regression (LogRegFusion), which provides a marginal improvement over ProbEn at the cost of training on a carefully curated multimodal dataset. Our analysis shows that learned fusion can be seen as a generalization of ProbEn that no longer assumes conditionally-independant modalities~\eqref{eq:logreg}.
}
\vspace{-2mm}
\begin{tabular}{l|c|c|c}
\hline
{\em Method} & {\em Day} & {\em Night} & {\em All} \\
\hline
RGB & 14.56 & 27.42 & 18.67 \\
Thermal & 24.59 & 7.76 & 18.99 \\ \hline
Pooling & 37.92 & 22.61 & 32.68 \\
NMS (MaxFusion) & 13.25 & 6.42 & 10.78 \\
AvgScore
& 21.68 & 15.16 & 19.53 \\
AvgLogits
& 18.78 & 11.70 & 16.28 \\
LogRegFusion & 10.70 & 6.11 & 9.08\\
ProbEn & 10.21 & {5.45} & {8.62} \\
\hline
ProbEn+bbox & 9.93 & 5.41 & 8.50 \\ 
\hline
\end{tabular}
\vspace{-3mm}
\label{tab:discussion-late-fusion}
\end{table}
}

{\bf Summing vs. averaging logits}.
Let us now revisit prior approaches to logit-based fusion in detail. Late fusion was popularized by video classification networks that made use of two-stream architectures~\cite{simonyan2014two}. This seminal work proposed an influential baseline for ``fusing softmax scores" by averaging. However, practical implementations average logits~\cite{two-stream-github-1,two-stream-github-3} or sum logits~\cite{two-stream-github-2}, often omitting the final softmax~\cite{wu2015modeling} because one can obtain a class prediction by simple maximization of the fused logits. In the classification setting, the distinction between summing versus averaging does not matter because both produce the same argmax label prediction. {\em But the distinction does matter in detection, which requires ranking and comparison of scores for non-maximal suppression (NMS) and global thresholding}. Intuitively, summing allows detections to become more confident as more modalities agree, while averaging does not. Most crucially, summing logits allows one to optimally compare detections with missing modalities, which is frequently needed in NMS whenever all modalities fail to fire on a given object. Here, optimality holds in the Bayesian sense whenever modalities are conditionally independent (as derived in \eqref{eq:sumfuse}).

{\bf Fusion from logits.} We can succintly compare various fusion approaches from the logit perspective with the following:
\begin{align}
    s_{\text{AvgLogit}}[k] &= .5(s_1[k] + s_2[k])\\
    s_{\text{Bayes}}[k] &=  s_1[k] + s_2[k]
\end{align}
It is easy to see that  $$s_{\text{AvgLogit}}[k] \leq s_{\text{Bayes}}[k]$$
Note that the relative ordering of the fused logits does {\em not} necessarily imply the same holds for the final posterior because the other class logits are needed to compute the softmax partition function. One particularly simple case to analyze is a single-class detector $k \in \{0,1\}$, as is true for the KAIST benchmark (that evaluates only pedestrians). 
Here we can analytically compute posteriors by looking at the {\em relative} logit score $s_i = s_i[1] - s_i[0]$ for modality $i$ (by relying on the well-known fact that a 2-class softmax function reduces to a sigmoid function of the relative input scores). We visualize the fused probability as a function of the relative per-modality logits $s_1$ and $s_2$ in Fig.~\ref{fig:landspaces}. Finally, Table~\ref{tab:discussion-late-fusion} explicitly compares the performance of such fusion approaches with other diagnostic variants. We refer the reader to both captions for more analysis.  

\begin{figure}[t]
\centering
\includegraphics[width=0.6\linewidth]{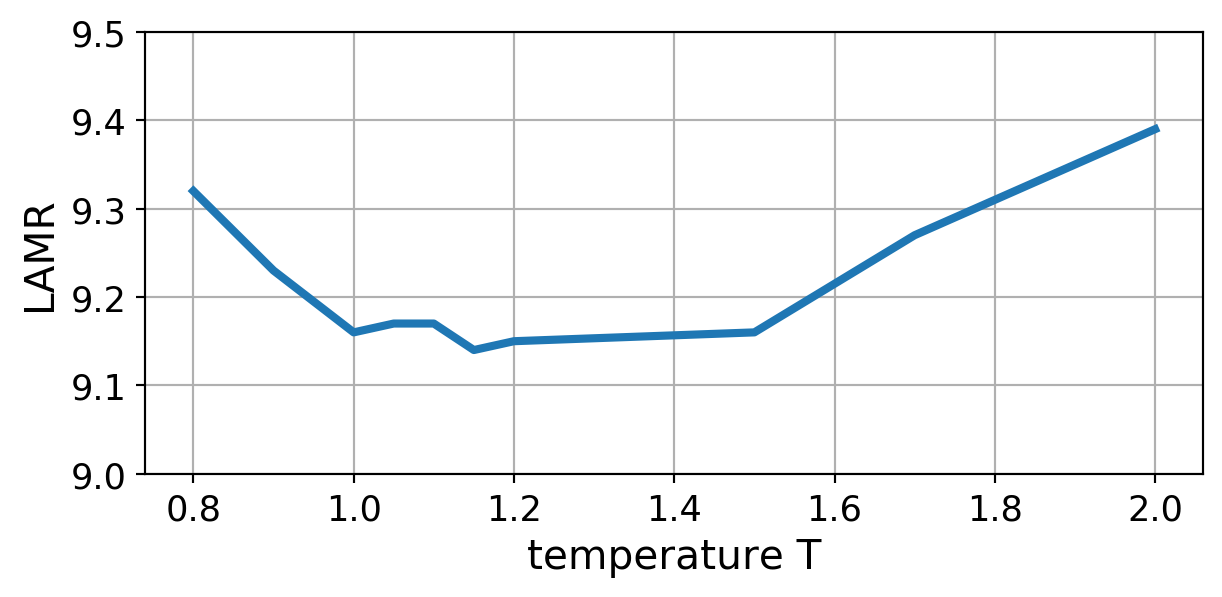}
\vspace{-3mm}
\caption{\small
{\bf LAMR as a function of a calibration temperature parameter $T$} (designed to return more realistic probabilities)~\cite{guo2017calibration} on KAIST reasonable-test,
We fuse detections from two single-modal detectors (RGB and thermal).
Here, $T$=1 corresponds to ProbEn. Tuning the temperature $T$ yields only marginally better performance. 
We conjecture that the scores from the two single-modal detectors are already comparable, presumably because both of them are trained with the same loss function, annotation labels, and network architecture.}
\vspace{-0mm}
\label{fig:calibration}
\end{figure}

{
\setlength{\tabcolsep}{0.35em} 
\begin{table}[t]
\scriptsize
\centering 
\caption{\small
\textbf{Late-fusion methods on different underlying detectors} measured by LAMR$\downarrow$ on KAIST reasonable-test. This table is comparable to Table 1 in the main paper. \emph{A}: RGB detector; \emph{B}: Thermal detector; \emph{C}: EarlyFusion detector; \emph{D}: MidFusion detector. 
Clearly, ProbEn consistently outperforms all other late-fusion methods. Interestingly, fusing detections from non-independent detectors (e.g., {\em A+B+D}) achieves better performance than independent detectors (e.g., {\em A+B}).
Lastly, probabilistically fusing boxes (using v-avg) improves further over 8 / 9 fusion methods.
}
\vspace{-2mm}
\begin{tabular}{l|l|c|c|c|c|c|c|c|c }
\hline
{\em Method} & {\em A+B} & {\em A+C} & {\em A+D} & {\em B+C} & {\em B+D} & {\em C+D} & {\em A+B+C} & {\em A+B+D} & {\em A+B+C+D} \\
\hline
Pooling & 32.68 & 28.87 & 29.70 & 36.68 & 36.36 & 23.24 & 43.04 & 43.56 & 46.03\\
AvgScore & 19.53 & 19.94 & 18.67 & 21.58 & 18.18 & 22.26 & 21.98 & 21.06 & 24.06\\ 
NMS & 10.85 & 11.59 & 13.05 & 18.74 & 13.81 & 14.18 & 10.91 & 12.11 & 12.09\\ 
ProbEn & 8.62 & \textbf{9.63} & 10.99 & 16.88 & 11.90 & 11.58 & 8.40 & 8.54 & 8.21\\
ProbEn + bbox & \textbf{8.50} & 9.87 & \textbf{10.30} & \textbf{16.87} & \textbf{11.20} & \textbf{11.32} & \textbf{8.55} & \textbf{7.66} & \textbf{7.45}\\
\hline
\end{tabular}
\vspace{-1mm}
\label{tab:summary-KAIST}
\end{table}
}

\begin{figure}[h!]
{\scriptsize \ {\bf (a)} RGB-detector \hspace{4mm} {\bf (b)} thermal-detector \hspace{6mm} {\bf (c)} NMS fusion \hspace{5mm} {\bf (d)} ProbEn fusion} \ \\
\centering 
\vspace{-3mm}
\includegraphics[width=1\linewidth, trim= 0cm 0cm 0cm 1.3cm, clip]{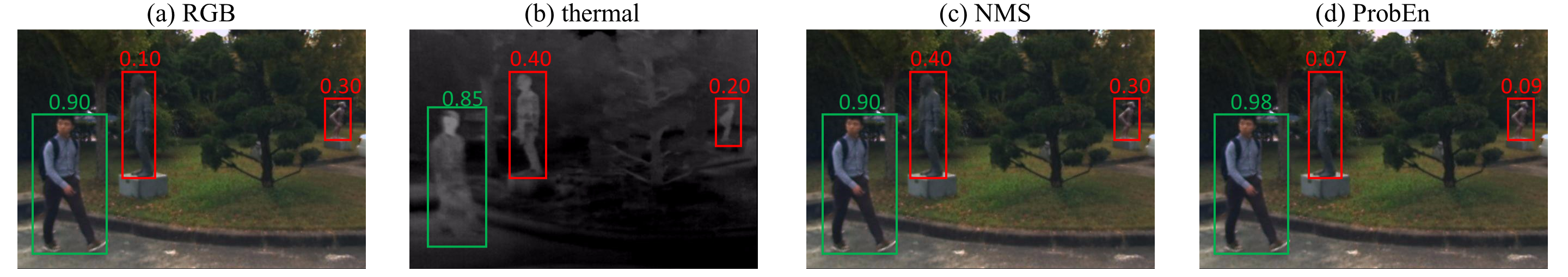}
\vspace{-7mm}
\caption{\small ProbEn handles false positives by lowering scores.
Fig.~\ref{fig:landspaces} (d) shows that ProbEn will {\em reduce} the fused score of overlapping detections with at least one low-scoring modality.
This is an example from KAIST, where RGB- and thermal-detectors produce {\color{purered}false-positive} pedestrian detections for the statues. NMS fusion keeps the higher-scoring {\color{purered}false-positive}, while ProbEn lowers the fused score while keeping the higher score for the {\color{darkgreen}true-positive} (that contain overlapping detections with consistently high scores).
}
\label{tab:handle-FPs}
\vspace{-2mm}
\end{figure}

\section{Score Calibration for Fusion}
\label{sec:score-calibration}
ProbEn assumes that detectors return true class posteriors. However, deep networks are notoriously over-confident in their predictions, even when wrong~\cite{guo2017calibration}.
One popular calibration strategy is adding a temperature parameter $T$ to the final softmax, typically to ``soften" overconfident estimates~\cite{guo2017calibration}.
This can be implemented by scaling logits by a temperature $T$:
\begin{equation}
s_i[k] \leftarrow s_i[k]/T, \qquad T > 0
\end{equation}
In the two-modality detection setting, because monotonic transformations of probability scores will not affect ranks (and hence not effect LAMR or AP), one can show that we need only calibrate one of two modalities. In practice, we calibrate thermal detector scores so as to better match scores from the RGB detector.
Figure~\ref{fig:calibration} plots LAMR as a function of a single scalar temperature $T$ used to scale thermal detections. Tuning $T$ yields only a marginal improvement over standard ProbEn (i.e., when $T=1$). 
We conjecture that the two single-modal detectors are trained with the same annotation and network architecture, making their output scores comparable to each other already.

\begin{figure}[t!]
\centering
\includegraphics[width=0.5\linewidth]{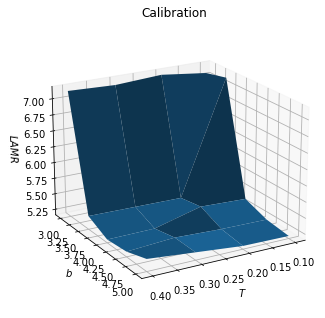}
\vspace{-7mm}
\caption{\small
{\bf LAMR as a function of calibration temperature parameter $T$ and shift parameter $b$}~\cite{guo2017calibration} on the KAIST validation set.
We fuse  single-modal detectors (RGB and thermal trained in-house) and an off-the-shelf detector GAFF~\cite{zhang2021guided}.
Clearly, both the temperature $T$ and shift $b$ greatly affect the final detection performance.} 
\vspace{-0mm}
\label{fig:Calib2D}
\end{figure}

Interestingly, when we ensemble an off-the-self multimodal detector GAFF~\cite{zhang2021guided}, our Thermal and RGB detectors (trained in-house), we find score calibration is particularly important. 
Importantly, we find that calibration requires not only a temperature variable but also a shift variable on the logits of GAFF. We conjecture that this is because GAFF is trained in a very different way; we do not know how GAFF is trained as there is not a publicly available codebase.
Fig.~\ref{fig:Calib2D} depicts the miss-rate as a function of the two variables, temperature $T$ and shift $b$. Clearly, the shift variable $b$ makes a significant impact on the fusion results.

{
\setlength{\tabcolsep}{0.87em} 
\begin{table}[t!]
\small
\centering
\caption{ \small
\textbf{Late-fusion methods on different underlying detectors} on FLIR dataset, measured by percent AP$\uparrow$ in percentage.  
\emph{A}: thermal detector; \emph{B}: EarlyFusion detector; \emph{C}: MidFusion detector. 
Our ProbEn method consistently outperforms other late-fusion methods. By fusing all the underlying detectors, ProbEn performs the best.
Lastly, probabilistically fusing boxes (using v-avg) improves further for 3 / 4 fusion methods.
}
\vspace{-2mm}
\begin{tabular}{l|l|c|c|c }
\hline
{\em Method} & {\em A+B} & {\em A+C}  & {\em B+C} & {\em A+B+C}\\
\hline
Pooling & 54.04 & 61.48 & 63.38 & 53.66 \\
AvgScore & 81.65 & 81.47 & 82.43 & 82.65\\ 
NMS & 81.75 & 82.34 & 82.43 & 83.14 \\ 
ProbEn & \textbf{82.05} & 82.26 & 82.67 & 83.27 \\
ProbEn + bbox & 81.93 & \textbf{82.85} & \textbf{83.04} & \textbf{83.76} \\
\hline
\end{tabular}
\label{tab:summary-FLIR}
\end{table}
}

\section{Further Study of Weighted Score Fusion}
\label{sec:weighted-fusion}
\begin{figure*}[t]
\centering 
\hspace{1mm}  {\small RGB frame} \hspace{9mm}  {\small thermal frame} \hspace{10mm}  {\small annotations} \hspace{9mm}  {\small our predictions} \hspace{1mm} \\
\centerline{
  \includegraphics[width=0.24\linewidth, trim= 0cm 4cm 0cm 0cm, clip]{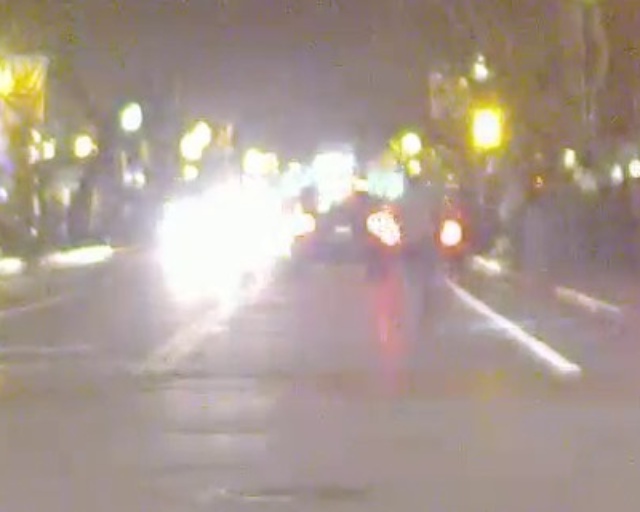}
  \includegraphics[width=0.24\linewidth, trim= 0cm 4cm 0cm 0cm, clip]{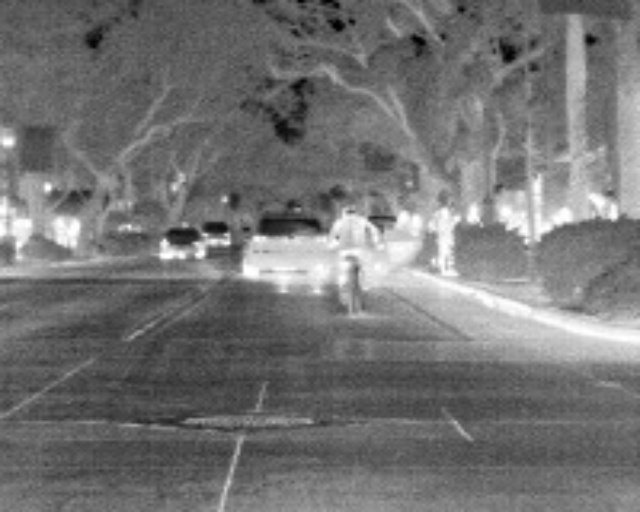}
  \includegraphics[width=0.24\linewidth, trim= 0cm 4cm 0cm 0cm, clip]{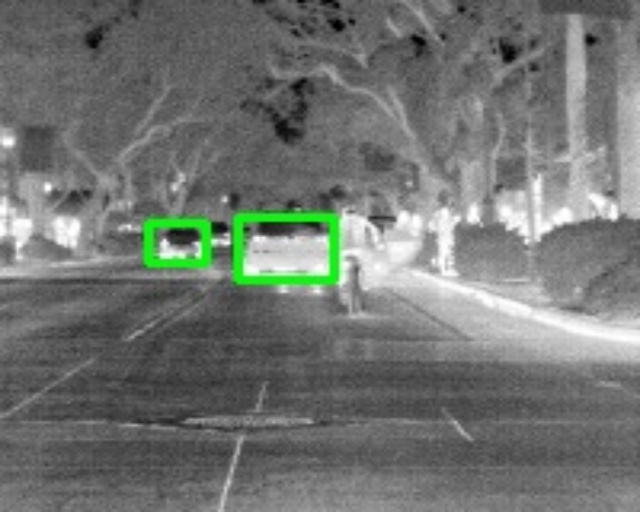}
  \includegraphics[width=0.24\linewidth, trim= 0cm 4cm 0cm 0cm, clip]{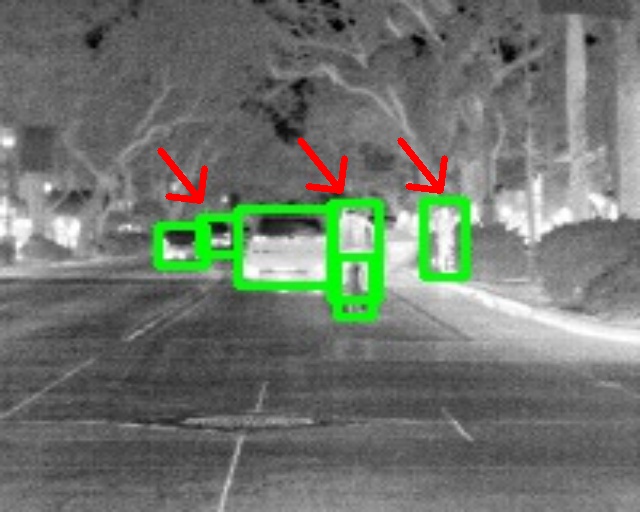}
}
\vspace{-4mm}
\caption{\small
We zoom in a frame from Fig.~\ref{fig:flir_bad_annotation} to visualize more clearly that the ground-truth anntoations can even miss {\tt bicycles} and {\tt persons} as shown in the third image. In contrast, our ProbEn model can detect these miss-labeled objects (cf. red arrows). This shows the issues in the FLIR dataset.
}
\vspace{-3mm}
\label{fig:flir_bad_annotation_zoom_in}
\end{figure*}

\begin{figure*}[t]
\centering 
\centerline{
  \includegraphics[width=0.24\linewidth, trim= 0cm 4cm 0cm 0cm, clip]{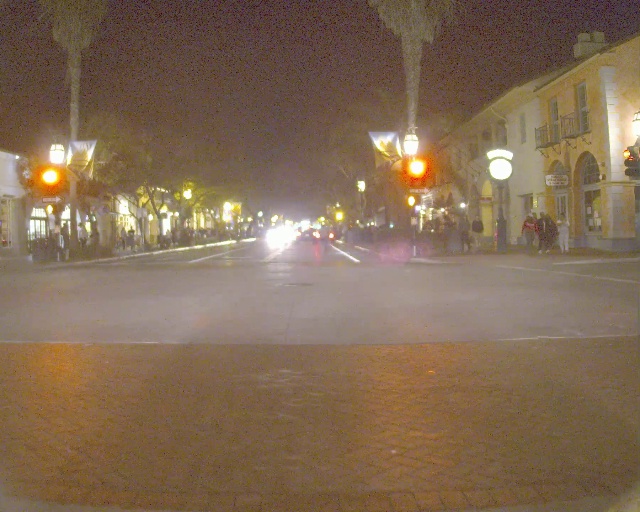}
  \includegraphics[width=0.24\linewidth, trim= 0cm 4cm 0cm 0cm, clip]{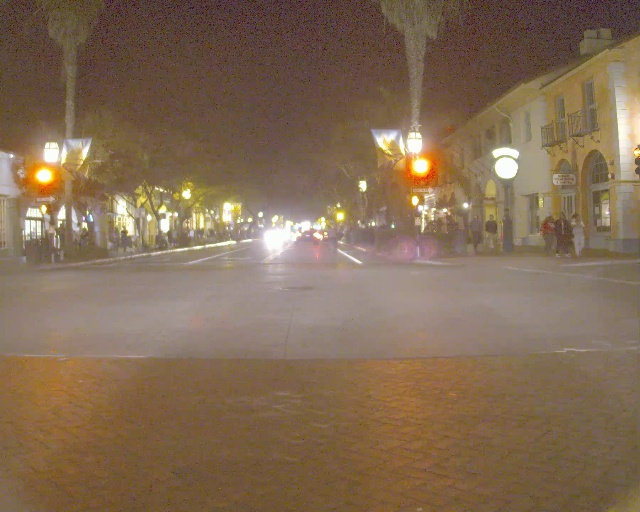}
  \includegraphics[width=0.24\linewidth, trim= 0cm 4cm 0cm 0cm, clip]{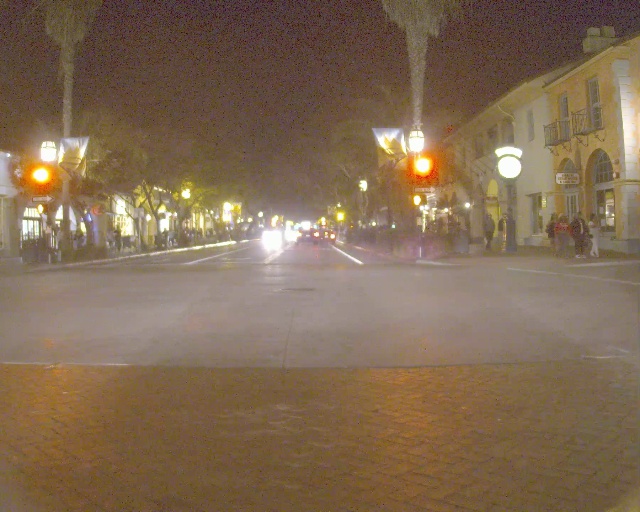}
  \includegraphics[width=0.24\linewidth, trim= 0cm 4cm 0cm 0cm, clip]{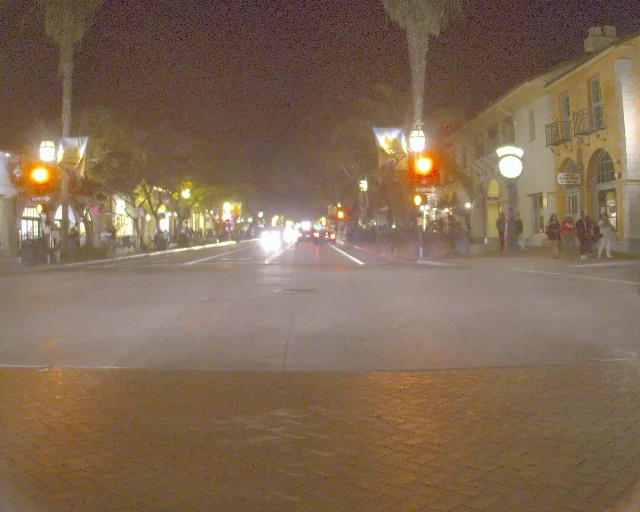}
}
\centerline{
  \includegraphics[width=0.24\linewidth, trim= 0cm 4cm 0cm 0cm, clip]{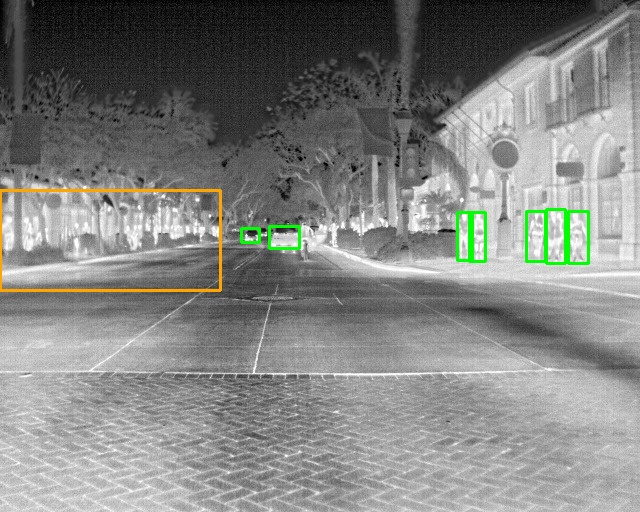}
  \includegraphics[width=0.24\linewidth, trim= 0cm 4cm 0cm 0cm, clip]{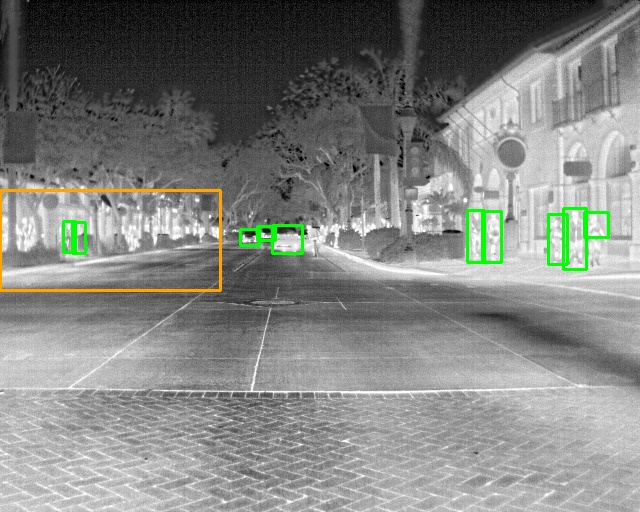}
  \includegraphics[width=0.24\linewidth, trim= 0cm 4cm 0cm 0cm, clip]{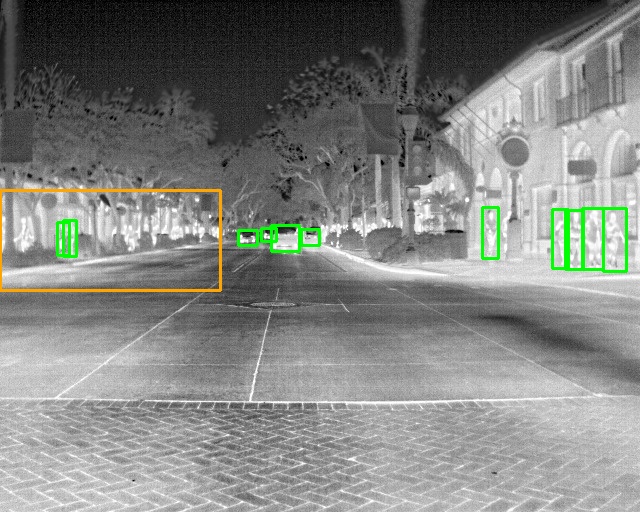}
  \includegraphics[width=0.24\linewidth, trim= 0cm 4cm 0cm 0cm, clip]{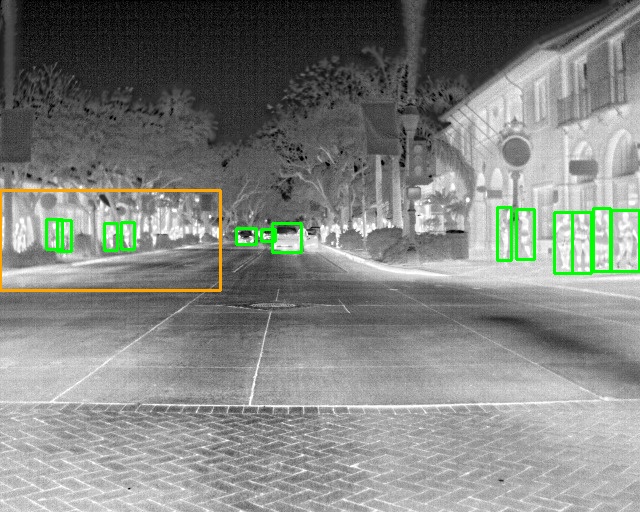}
}
\centerline{
  \includegraphics[width=0.24\linewidth, trim= 0cm 4cm 0cm 0cm, clip]{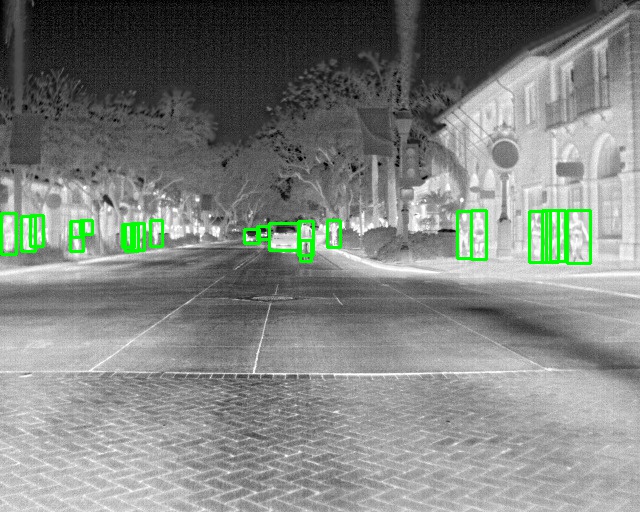}
  \includegraphics[width=0.24\linewidth, trim= 0cm 4cm 0cm 0cm, clip]{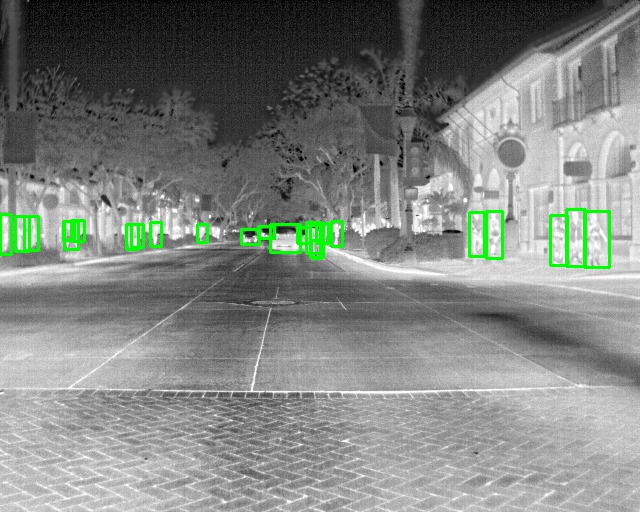}
  \includegraphics[width=0.24\linewidth, trim= 0cm 4cm 0cm 0cm, clip]{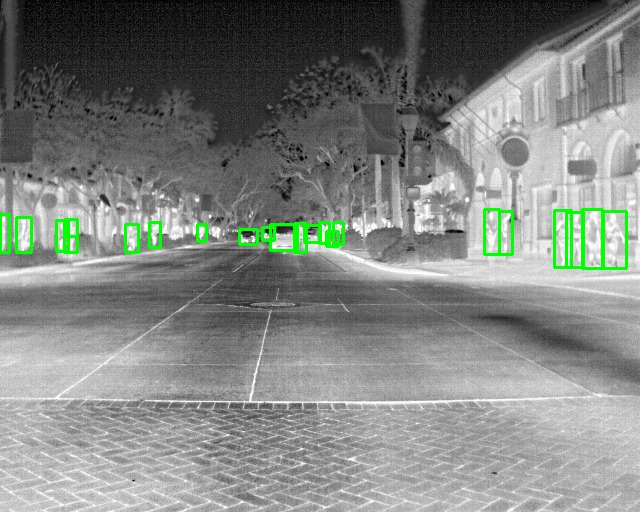}
  \includegraphics[width=0.24\linewidth, trim= 0cm 4cm 0cm 0cm, clip]{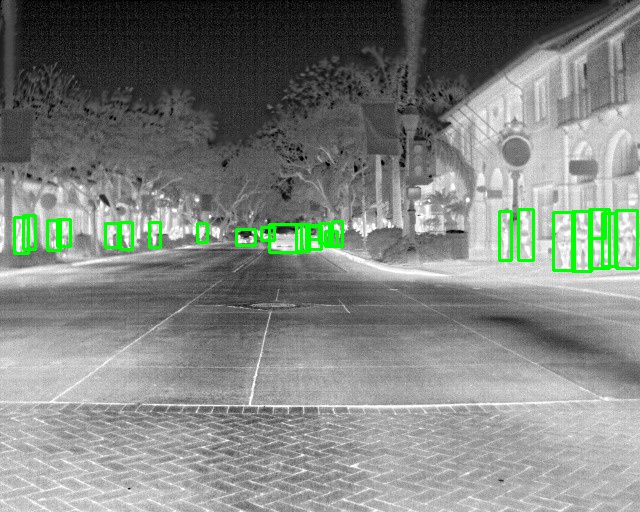}
}
\vspace{-3mm}
\caption{\small
We demosntrate inconsistent annotations in FLIR dataset with four consecutive frames in the validation set.
{\bf top-row} lists four RGB frames for reference.
{\bf mid-row} displays thermal images and the ground-truth annotations. Looking at the annotations in the orange rectangle, we can see that the annotations are not consistent across frames. This is an critical issue that prevent learning-based late fusion from improving further on the FLIR dataset.
{\bf Bottom-row} displays the detection results by ProbEn of the three models (Thermal, Early, and Mid). Interestingly, the predictions look more reasonable in detecting pedestrians within the orange rectangles. In this sense, predictions is ``better'' than annotations, intuitively explaining why learning based late fusion does not improve performance further.
Please also refer to Fig.~\ref{fig:flir_bad_annotation_zoom_in} for a zoom-in visualization.
}
\label{fig:flir_bad_annotation}
\end{figure*}

All late fusion approaches discussed thus far do not require training on multimodal data. Because prior work on late fusion has also explored learned variants, we also consider 
(learned) linear combinations of single-modal logits:
\begin{align}
    s_{\text{Learned}}[k] &=  w_1[k] s_1[k] + w_2[k] s_2[k] \label{eq:logreg}
\end{align}
One can view ProbEn, AvgLogits, and Temperature Scaling as special cases of the above. ProbEn and AvgLogits use predefined weights that do not require learning and so are easy to implement. Temperature scaling requires single-modal validation data to tune each temperature parameter, but does not require multimodal learning. This can be advantageous in settings where modalities do not align (e.g., FLIR) or where there exists larger collections of single-modal training data (e.g., COCO training data for RGB detectors).
Truly joint learning of weights requires multimodal training data, but joint learning may better deal with correlated modalities by downweighting the contribution of modalities that are highly correlated (and don't provide independant sources of information). We experimented with joint learning of the weights with logistic regression. To do so, we assembled training examples of overlapping single-modal detections (and cached logit scores) encountered during NMS, assigning a binary target label (corresponding to true vs false positive detection). After training on such data, we observe a small improvement over non-learned fusion (Table~\ref{tab:discussion-late-fusion}), consistent with prior art on late fusion~\cite{simonyan2014two}. 
We also tested learning-based late fusion methods on the FLIR dataset. We further tested learning class priors. However,  these methods do not yield better performance than the simple non-learned ProbEn (both achieve 82.91 AP). The reason is that  FLIR annotations are inconsistent across frames, making it hard for learning-based late fusion methods to shine, as explained in Fig.~\ref{fig:flir_bad_annotation} and \ref{fig:flir_bad_annotation_zoom_in}.

\section{Further Study of Class Prior in ProbEn}
\label{sec:class-prior}
\begin{figure}
    \centering
    \includegraphics[width=.7\linewidth]{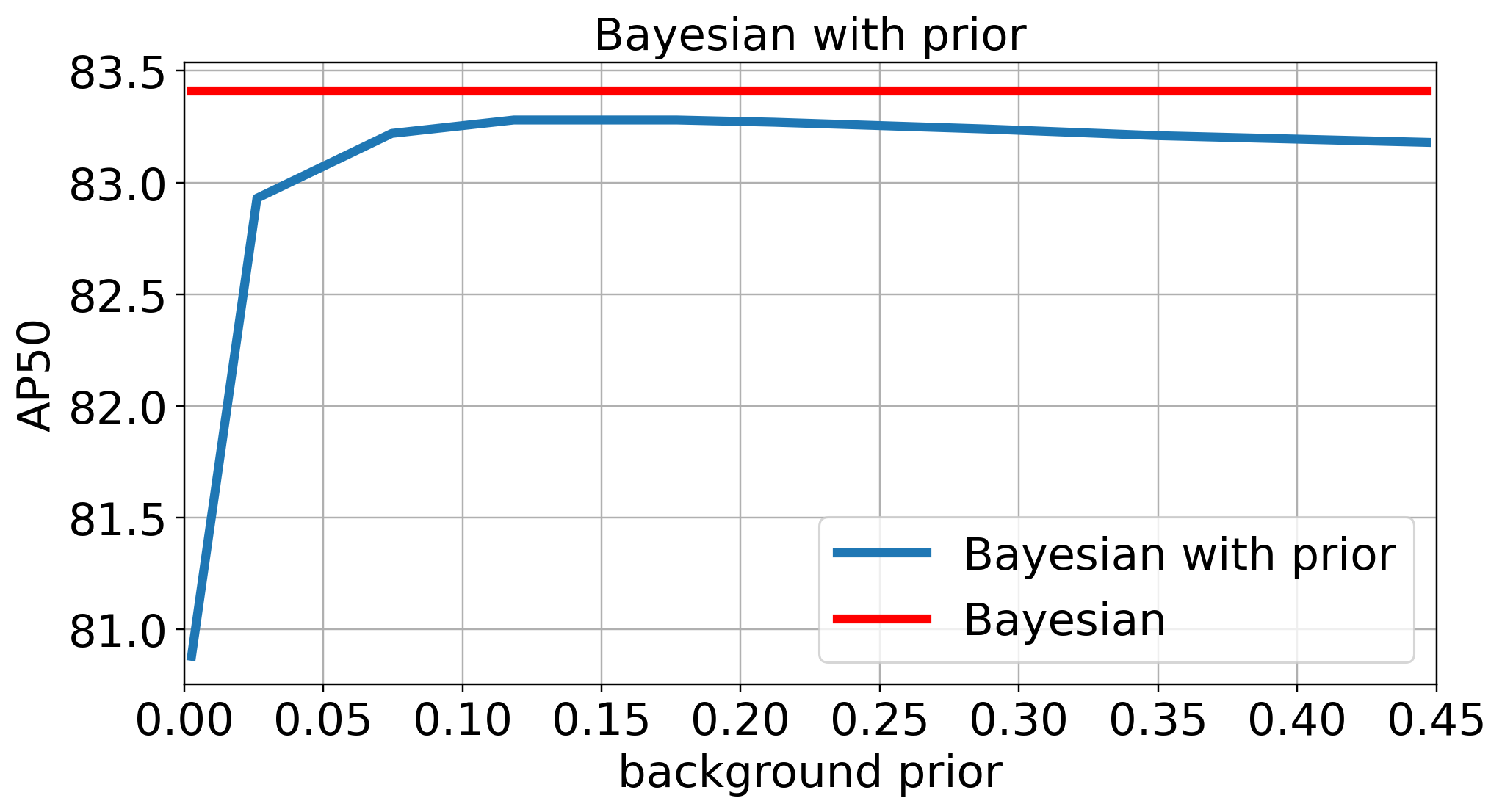}
    \vspace{-2mm}
    \caption{\small
    A study of ProbEn with class priors as class frequencies in the training set.
    We use FLIR dataset for this study as it has 3 imbalanced classes. We fuse three models (Thermal, Early and Mid) as used in the main paper.
    As there is a background class, we vary the background class and proportionally change the class priors.
    Clearly, ProbEn with uniform class priors performs better than using the computed priors. Tuning the background prior does not notably affect the final detection performance once this prior is set to be larger than 0.1.
    }
    \label{fig:backgroundprior}
\end{figure}

In the main paper, we assume uniform class priors when using ProbEn. Now we test ProbEn with computed class priors. For consistent experiments as done in the main paper, we use FLIR dataset and fuse three models (Thermal, Early and Mid). Recall that FLIR has imbalanced classes:  
{\tt person} (21,744), {\tt bicycle} (3,806), and {\tt car} (39,372).
First, we count the number of annotated objects of each of the three class, and assign the fourth background class with a dummy number. Then, we normalize them to be sum-to-one as class priors. We vary the background prior and evaluate the final detection performance measured by AP at IoU$>$0.5, as shown in Fig.~\ref{fig:backgroundprior}.
Clearly, ProbEn works better with uniform priors than the computed the class piriors.

\begin{figure}
    \centering
    \includegraphics[width=.7\linewidth]{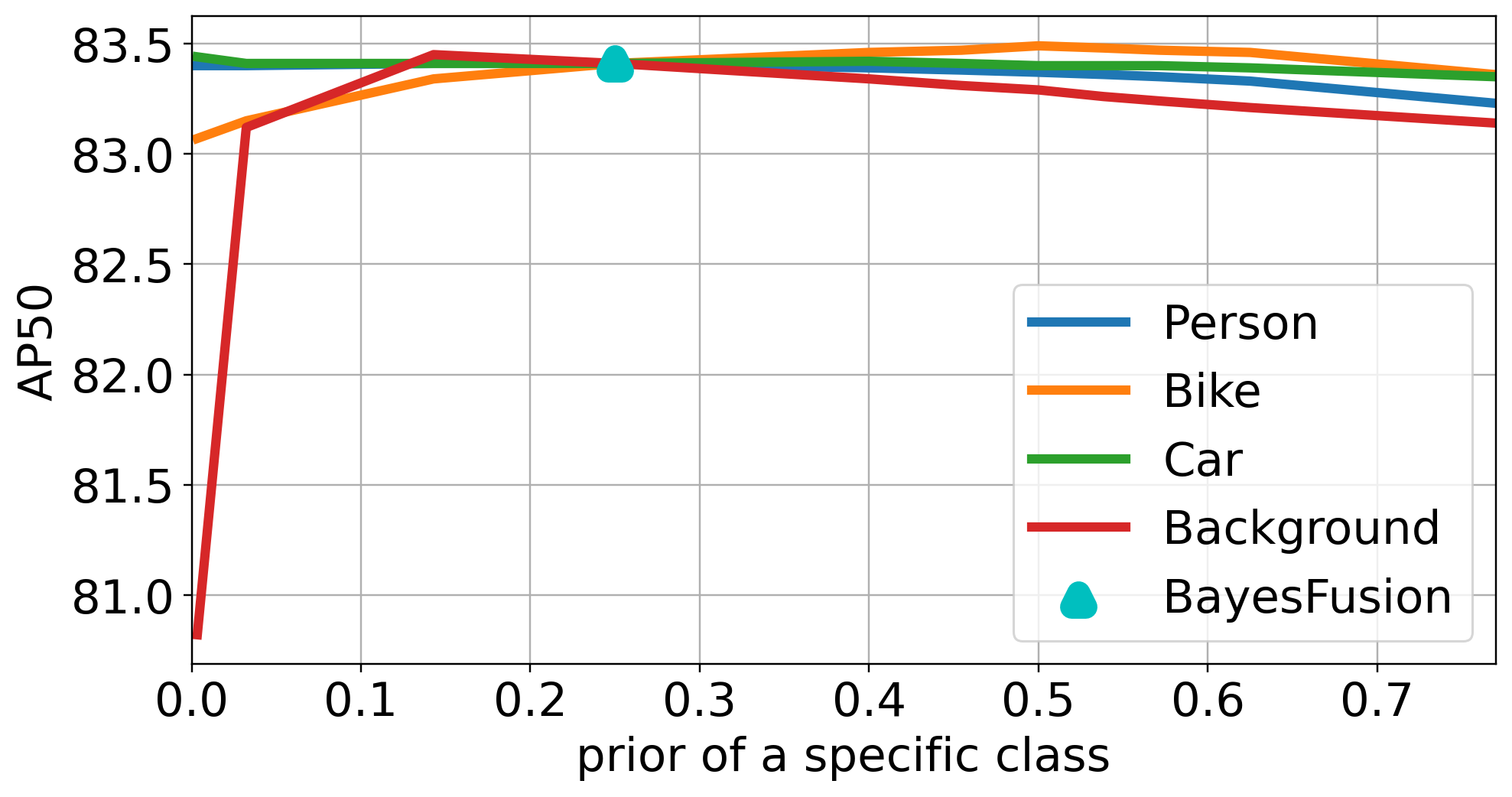}
    \vspace{-3mm}
    \caption{\small
    A study of tuning a single class prior while keeping others  the same. Motivated by the superior performance of ProbEn with uniform priors, we tune each of the class prior by fixing others the same. 
    We study this on the FLIR dataset by fusing three models (Thermal, Early and Mid).
    We can see that tuning specific classes only marginally improves detection performance. 
    }
    \label{fig:vary_per_class_prior}
\end{figure}

We further ablate which class is more important by manually assigning a prior. Concretely, we vary one class prior by fixing the others to be the same. We plot the performance vs. the per-class prior in Fig.~\ref{fig:vary_per_class_prior}. Tuning specific class priors yields marginal improvements compared to using uniform prior.

\section{A Detailed Derivation of Probabilistic Box Fusion}
\label{sec:box-fusion}

In the main paper, we present a probabilistic method to fuse multiple bounding boxes. Below is a detailed derivation. We write $\z$ for the continuous random variable defining the bounding box (parameterized by its centroid, width, and height) associated with a given detection. We assume single-modal detections provide a posterior $p(\z|x_i)$ that takes the form of a Gaussian with a single variance $\sigma_i^2$, i.e., $p(\z|x_i)={\cal N}(\muu_i, \sigma_i^2\I)$
where $\muu_i$ are box coordinates predicted from modality $i$. We also assume a uniform prior on $p(\z)$, implying box coordinates can lie anywhere in the image plane. Doing so, we derive probabilistic box fusion:  
\begin{equation}
\begin{split}\small
& \ p(\z | x_1, x_2) \propto  \ p(\z | x_1) p(\z|x_2) \\
\propto & \ \exp\big(\frac{\Vert \z-\muu_1 \Vert^2}{-2\sigma_1^2} \big) \exp\big(\frac{\Vert \z-\muu_2 \Vert^2}{-2\sigma_2^2} \big) \\
\propto & \ \exp\big(\frac{\z^T\z-2\muu_1^T\z+\muu_1^T\muu_1}{-2\sigma_1^2} \big) \exp\big(\frac{\z^T\z-2\muu_2^T\z+\muu_2^T\muu_2}{-2\sigma_2^2} \big) \\
\propto & \ \exp\big(\frac{\z^T\z-2\muu_1^T\z+\muu_1^T\muu_1}{-2\sigma_1^2} +\frac{\z^T\z-2\muu_2^T\z+\muu_2^T\muu_2}{-2\sigma_2^2} \big) \\
\propto & \ \exp\Big(\frac{\frac{1}{\sigma_1^2} + \frac{1}{\sigma_2^2}}{-2} * ( \z^T \z -2 \frac{\frac{\muu_1^T}{\sigma_1^2} + \frac{\muu_2^T}{\sigma_2^2}}{\frac{1}{\sigma_1^2} + \frac{1}{\sigma_2^2}} * \z)\Big) \\
\propto & \ \exp(\frac{\frac{1}{\sigma_1^2} + \frac{1}{\sigma_2^2}}{-2} * ||\z-\muu||^2), \quad \text{where} \quad \muu=\frac{(\muu_1/\sigma_1^2 + \muu_2/\sigma_2^2 )}{(1/\sigma_1^2 + 1/\sigma_2^2)}
\end{split}
\nonumber
\end{equation}

\section{A Study of Fusing More and Better Models}
\label{sec:quantitative-results}

We study late fusion methods on more combinations of underlying detectors.
Table~\ref{tab:summary-KAIST},  \ref{tab:summary-FLIR} and \ref{tab:fuseSOTA} list results on KAIST and FLIR datasets, respectively.
Importantly, ProbEn consistently performs the best on each of combinations. Interestingly, applying ProbEn method to detectors that are not independent to each other (e.g., Thermal and MidFusion) can achieve better performance.
Admittedly, the improvements may not be statistically significant and overfitting may be an issue. This can not be resolved or studied further using contemporary datasets which are relatively small.
Therefore, we solicit a larger-scale dataset to benchmark multimodal detection in the commmunity.

{
\setlength{\tabcolsep}{1.7em} 
\begin{table}[t]
\centering
\small
\caption{ \small
\textbf{ProbEn always
outperforms NMS when applied to the same ensemble of (even strong) detections.}
Results are comparable to Table~\ref{tab:ablation_KAIST}.
}
\vspace{-3.5mm}
\begin{tabular}{lccc }
\hline
\multicolumn{4}{c}{\tt fusing MLPD and GAFF on KAIST (LAMR$\downarrow$ in \%)}\\
\hline
{\em method} 
& {\em Day} 
& {\em Night} 
& \cellcolor{lightlightgrey}{\em All} \\
\hline
MLPD     & 7.96 & 6.95 & \cellcolor{lightlightgrey}7.58  \\ 
GAFF     & 8.25 & {3.46} & \cellcolor{lightlightgrey}6.38  \\  
\hline
NMS (MLPD+GAFF)      & 7.63 & 6.76 & \cellcolor{lightlightgrey}7.24  \\ 
ProbEn (MLPD+GAFF)   & 6.23 & 3.79 & \cellcolor{lightlightgrey}{\bf 5.38} \\ 
\hline
NMS$_3$ w/ MLPD      & 7.34 & 7.03 & \cellcolor{lightlightgrey}7.13  \\
ProbEn$_3$ w/ MLPD   & 7.81 & 5.02 & \cellcolor{lightlightgrey}{\bf 6.76} \\
\hline
NMS$_3$ w/ GAFF      & 8.29 & {3.46} & \cellcolor{lightlightgrey}6.36  \\ 
ProbEn$_3$ w/ GAFF   & {6.04} & 3.59 & \cellcolor{lightlightgrey}{\bf 5.14} \\
\hline
\end{tabular}
\label{tab:fuseSOTA}
\vspace{-3.5mm}
\end{table}
}

\section{Qualitative Results and Video Demo}
\label{sec:visual-results}
\begin{figure}
\centering
\includegraphics[width=0.55\linewidth]{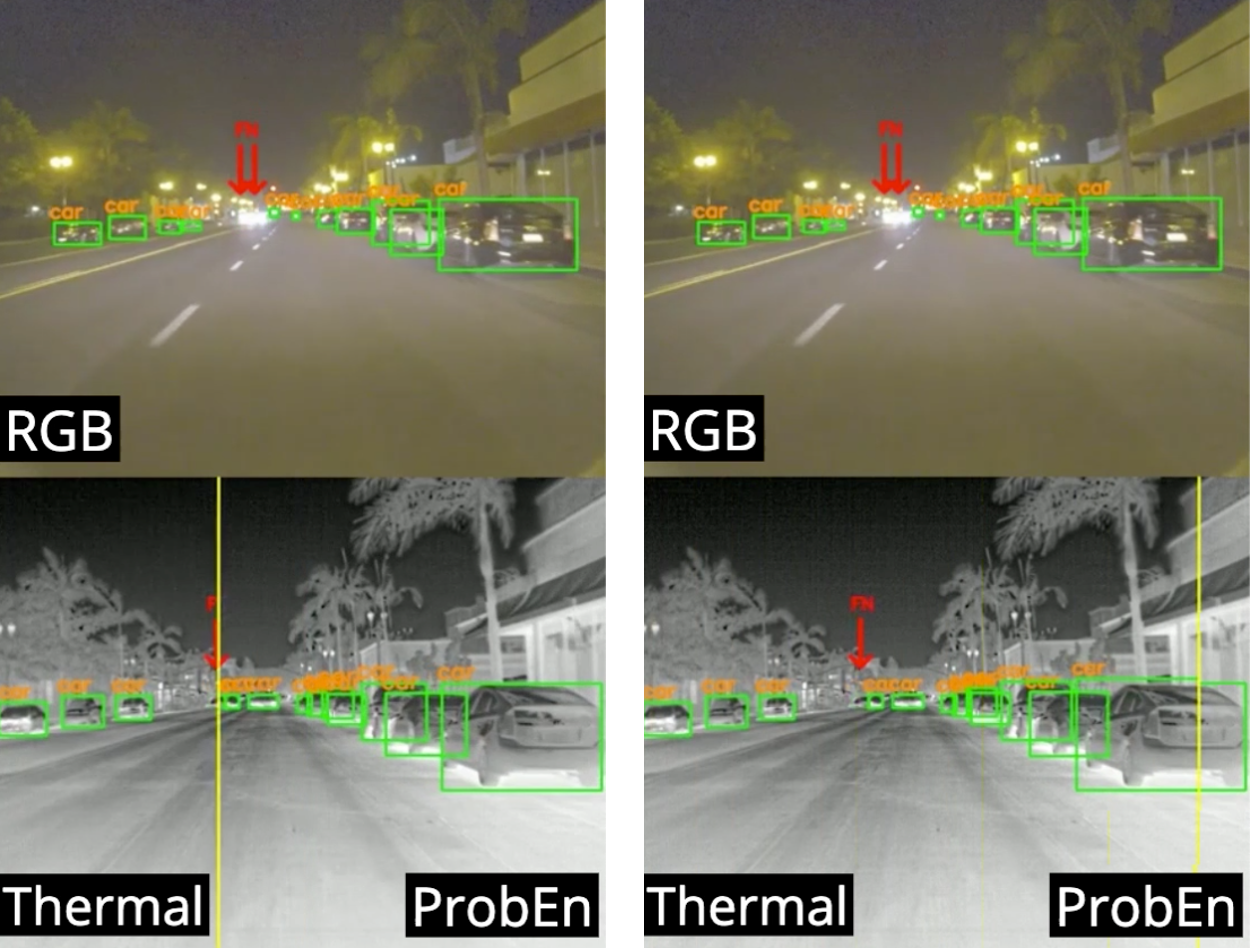}
\vspace{-2mm}
\caption{\small
We attach a demo video in our \href{https://github.com/Jamie725/RGBT-detection}{Github repository}.
The demo video is generated based on a testing video (captured at night) provided by the FLIR dataset.
Hereby we display two video frames for a same scene that compare detections by a thermal-only single-modal detector and the ProbEn method that fuses three detectors (Thermal, Early-fusion and Mid-fusion).
We can see Thermal detector mis-detects a car and produces larger bounding box for the rightmost car (right frame), in contrast, ProbEn successfully detects all the cars and produces tight bounding boxes.
We refer the reader to the video demo for convincing visualization.
}
\label{fig:video-demo-frames}
\end{figure}

In our \href{https://github.com/Jamie725/RGBT-detection}{github repository}, we attach a demo video on a testing video (captured at night) provided by the FLIR dataset. In the video, we compare the detection results by the Thermal model and ProbEn that fuses results of three models (Thermal + Eary + Mid). Recall that the FLIR dataset does not align RGB and thermal frames, and annotates only thermal frames. Therefore, we only provide RGB frames as reference (cf. Fig.~\ref{fig:video-demo-frames}).

\begin{figure*}[t]
\centering 
  \centerline{
      \includegraphics[width=0.21\linewidth]{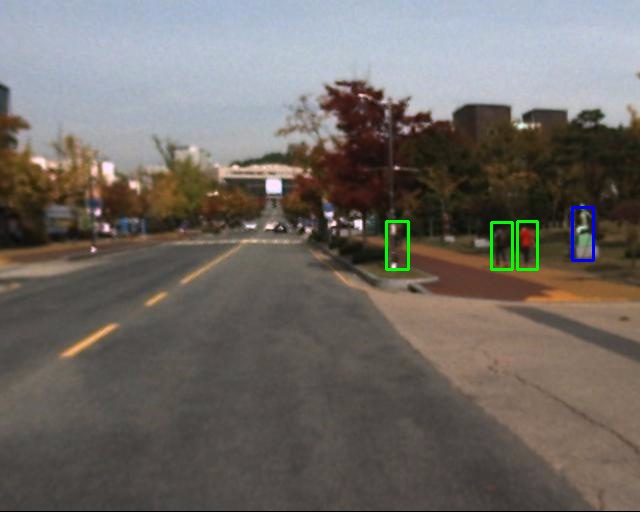}
      \includegraphics[width=0.21\linewidth]{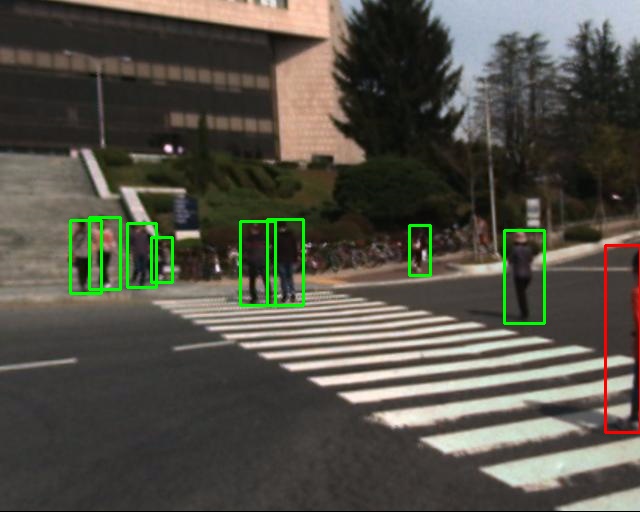}
      \includegraphics[width=0.21\linewidth]{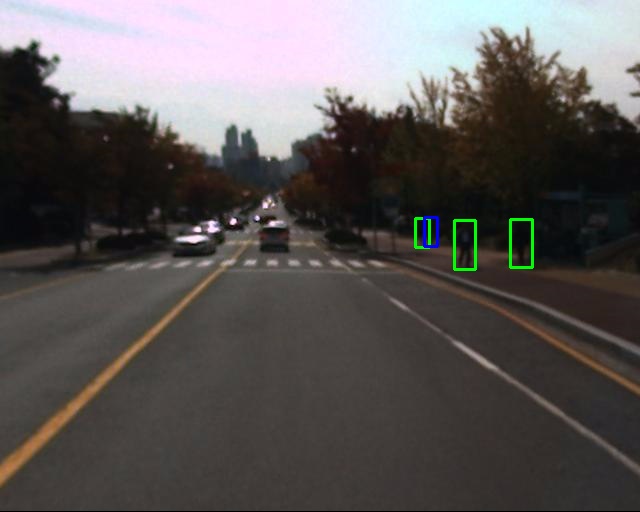}
      \includegraphics[width=0.21\linewidth]{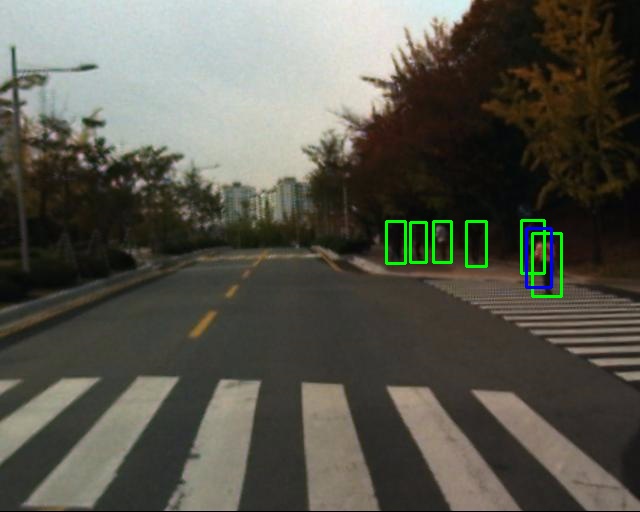}
  }
  \centerline{
      \includegraphics[width=0.21\linewidth]{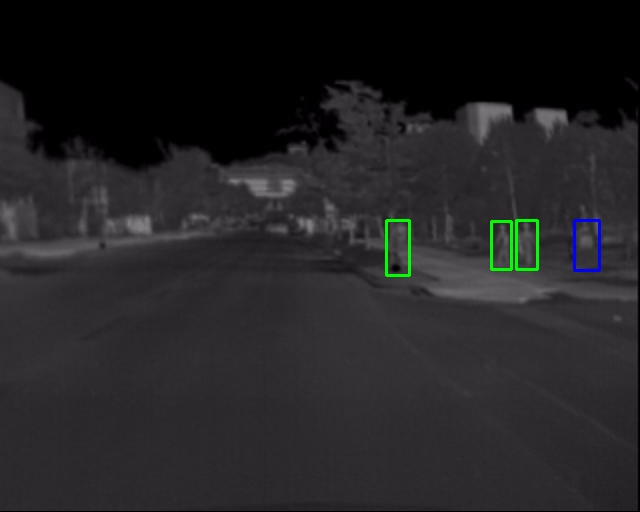}
      \includegraphics[width=0.21\linewidth]{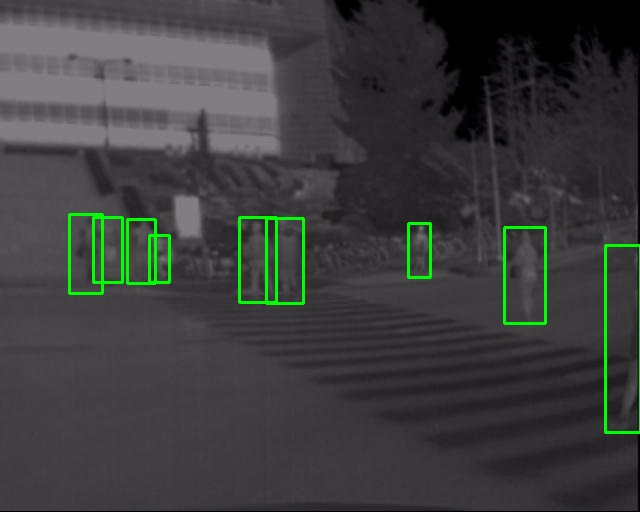}
      \includegraphics[width=0.21\linewidth]{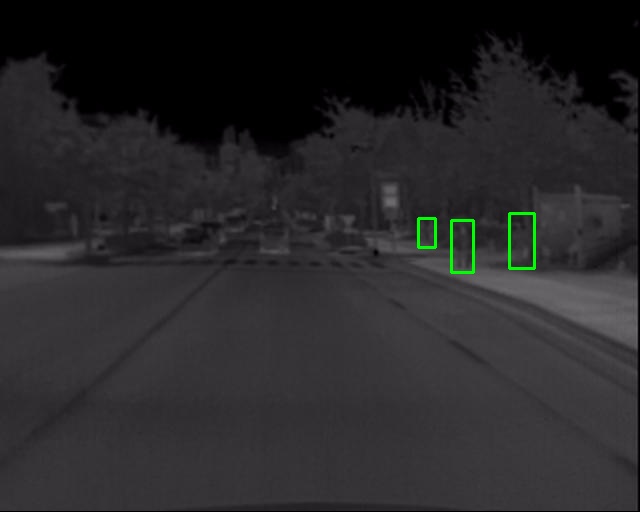}
      \includegraphics[width=0.21\linewidth]{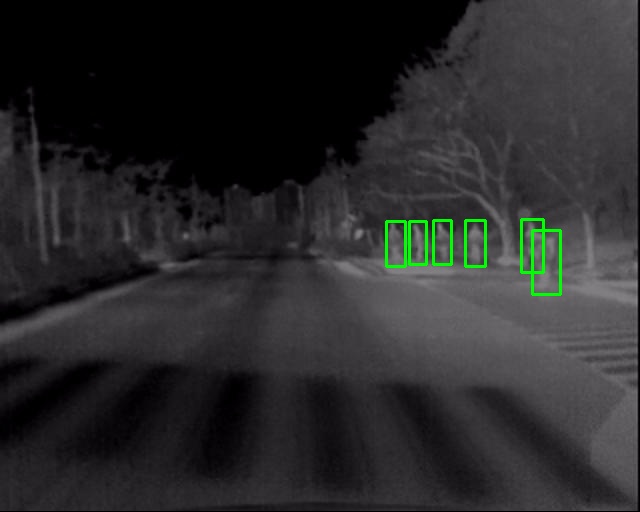}
  }
  \noindent\makebox[\linewidth]{\rule{0.8\paperwidth}{1pt}} 
  \vfill \bigskip
  \centerline{
      \includegraphics[width=0.21\linewidth]{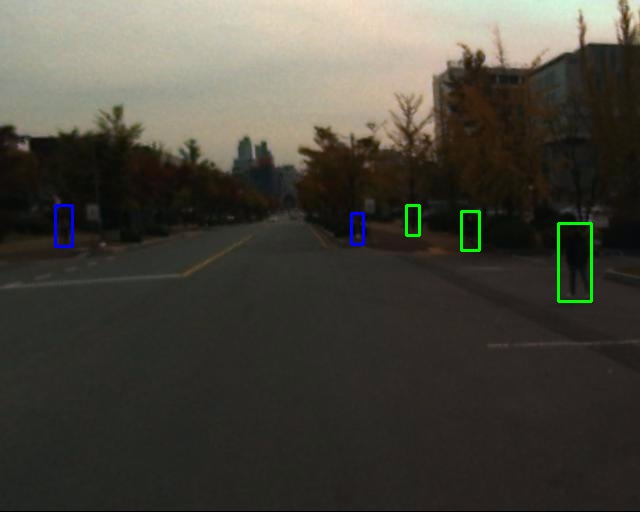}
      \includegraphics[width=0.21\linewidth]{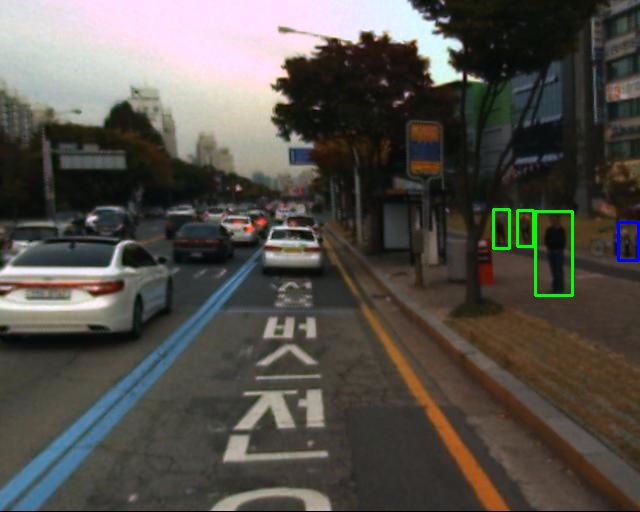}
      \includegraphics[width=0.21\linewidth]{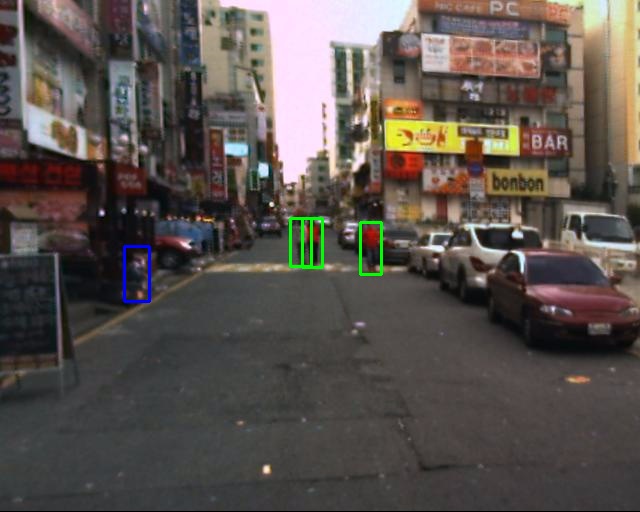}
      \includegraphics[width=0.21\linewidth]{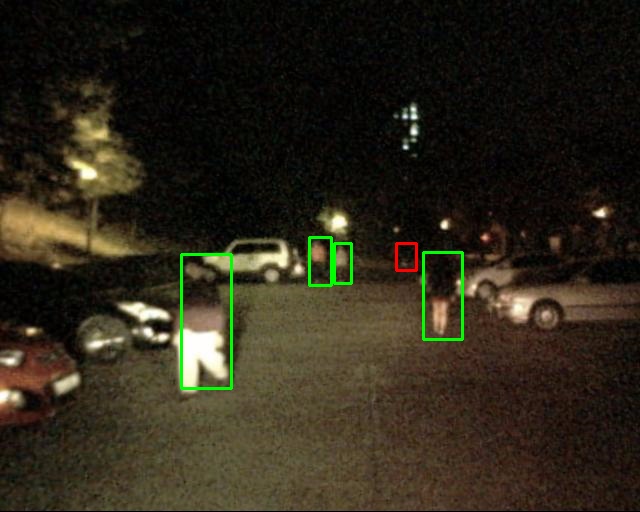}
  }
  \centerline{
      \includegraphics[width=0.21\linewidth]{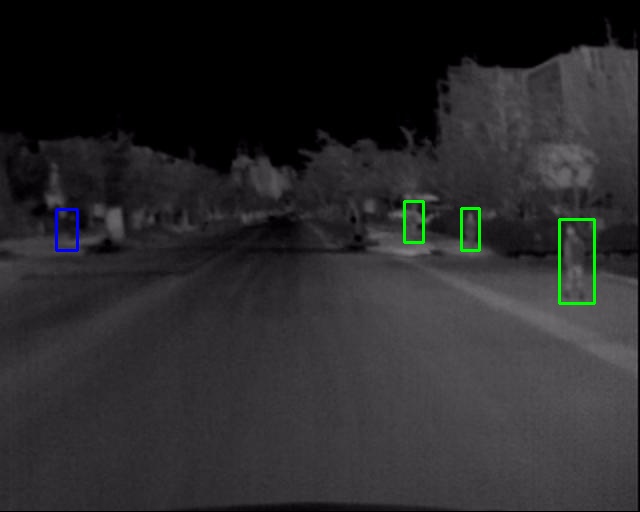}
      \includegraphics[width=0.21\linewidth]{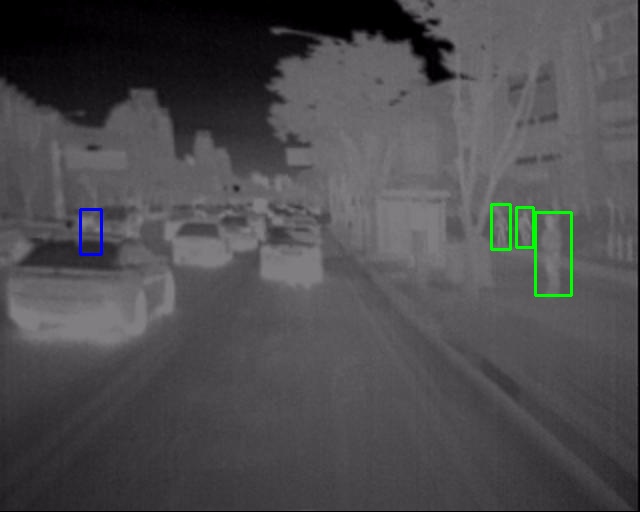}
      \includegraphics[width=0.21\linewidth]{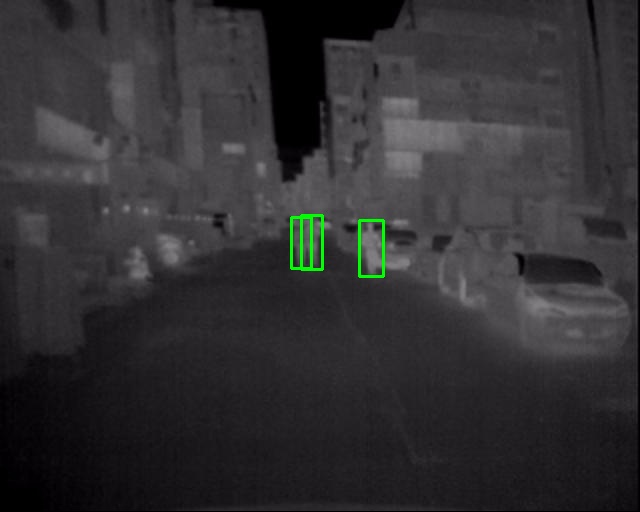}
      \includegraphics[width=0.21\linewidth]{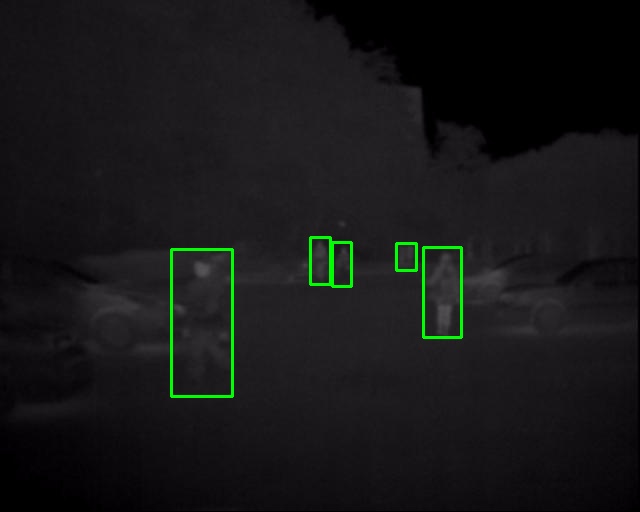}
  }
  \noindent\makebox[\linewidth]{\rule{0.8\paperwidth}{1pt}} 
  \vfill \bigskip
  \centerline{
      \includegraphics[width=0.21\linewidth]{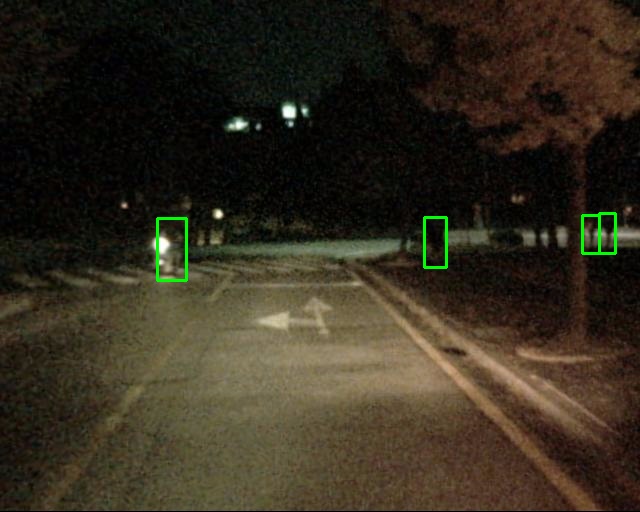}
      \includegraphics[width=0.21\linewidth]{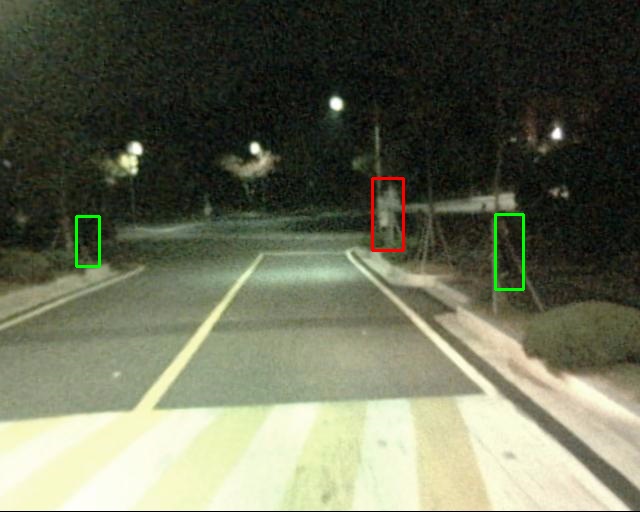}
      \includegraphics[width=0.21\linewidth]{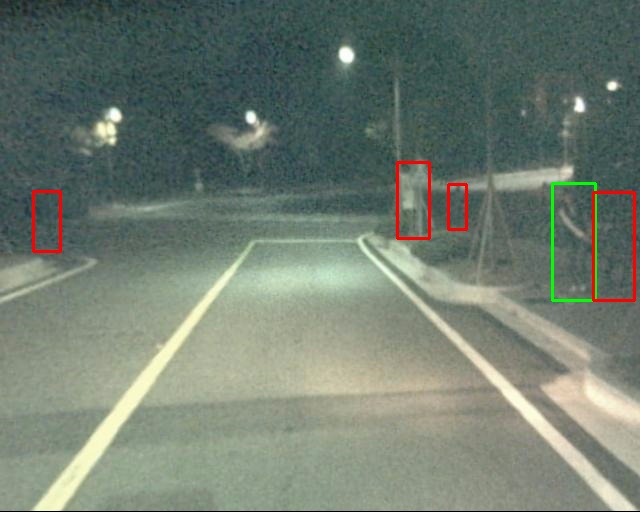}
      \includegraphics[width=0.21\linewidth]{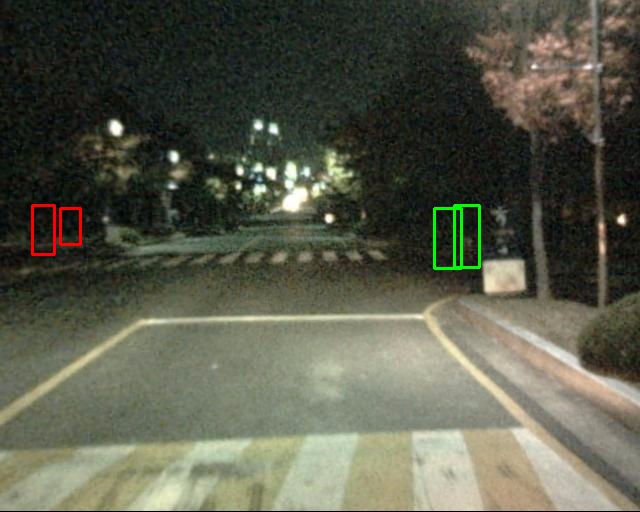}
  }
  \centerline{
      \includegraphics[width=0.21\linewidth]{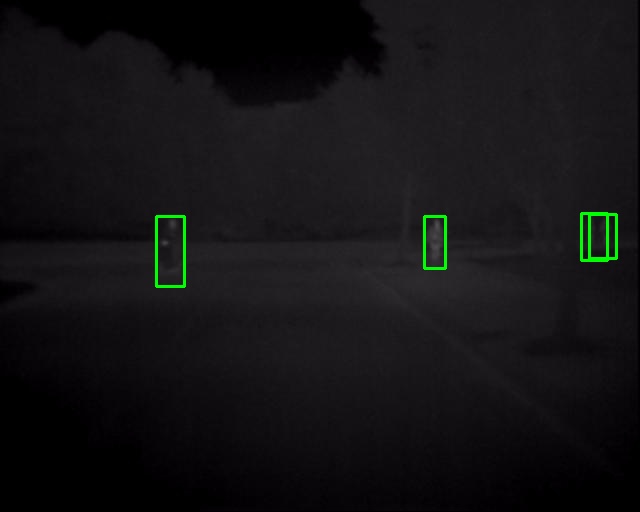}
      \includegraphics[width=0.21\linewidth]{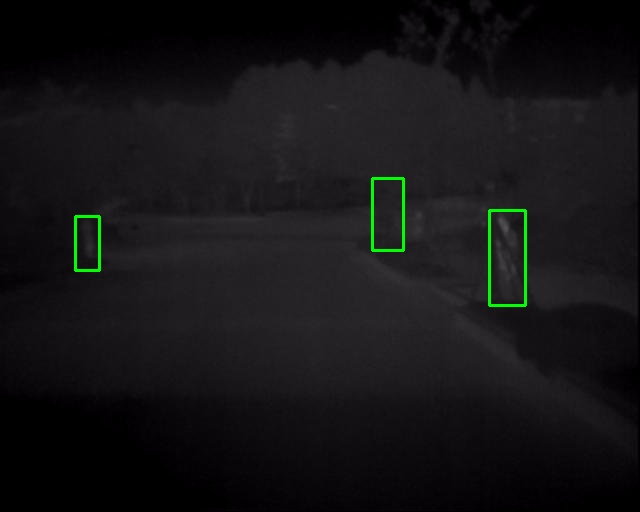}
      \includegraphics[width=0.21\linewidth]{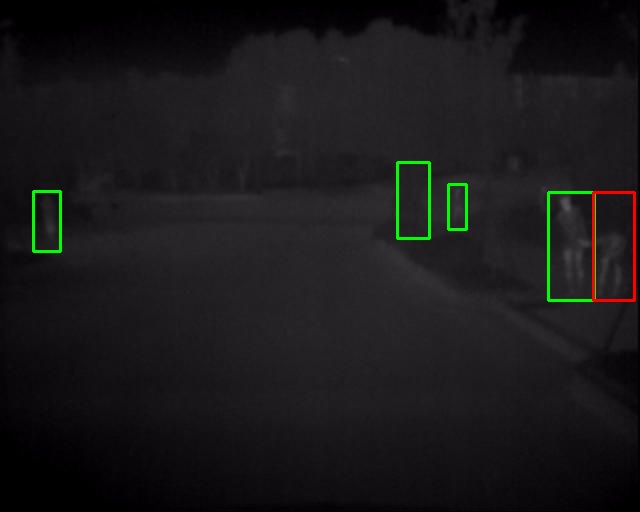}
      \includegraphics[width=0.21\linewidth]{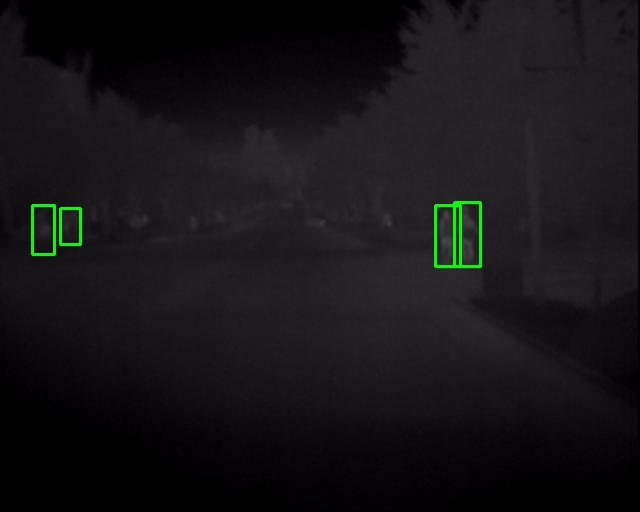}
  }
\vspace{-3mm}
\caption{\small 
Qualitative results on more testing examples in KAIST dataset.
We place RGB-thermal images in pairs: in each macro row, we show RGB images in the upper row and thermal images in lower row.
Over RGB images, we overlay the detection results from our MidFusion model; on the thermal images, we show results from our best-performing ProbEn model. 
{\color{darkgreen} Green}, {\color{red} red} and {\color{darkblue} blue} boxes stand for {\color{darkgreen} true positives}, {\color{red} false negative} (miss-detected persons) and {\color{darkblue} false positives}. 
}
\vspace{-4mm}
\label{fig:more-visual-KAIST}
\end{figure*}

\begin{figure*}[t]
\centering 
  \centerline{
      \includegraphics[width=0.22\linewidth]{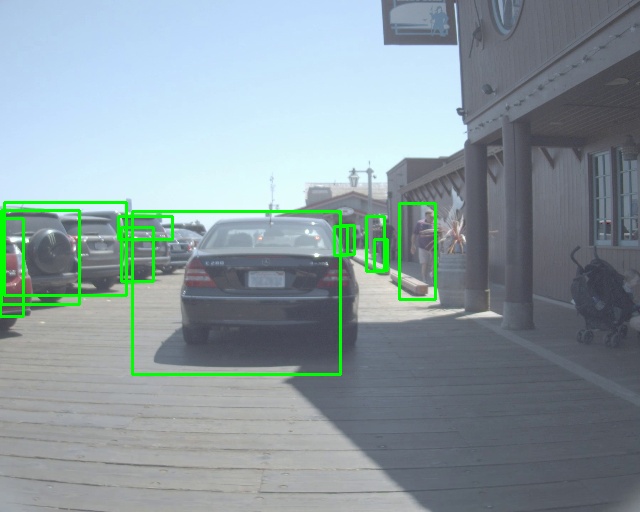}
      \includegraphics[width=0.22\linewidth]{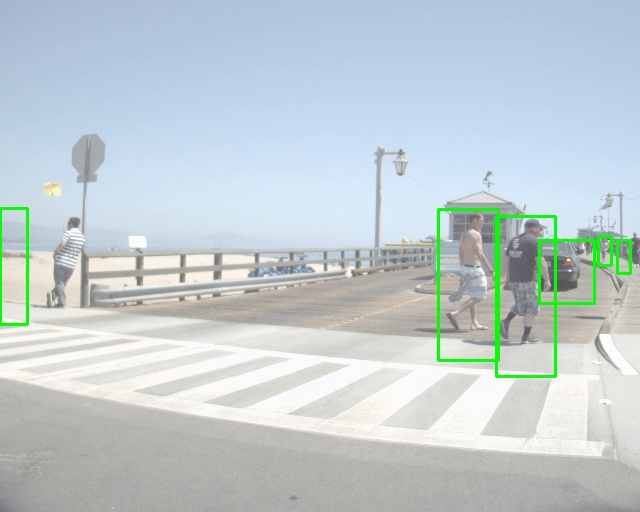}
      \includegraphics[width=0.22\linewidth]{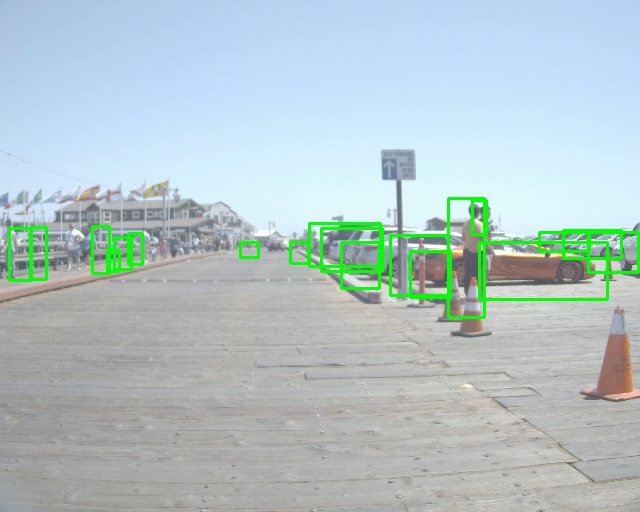}
      \includegraphics[width=0.22\linewidth]{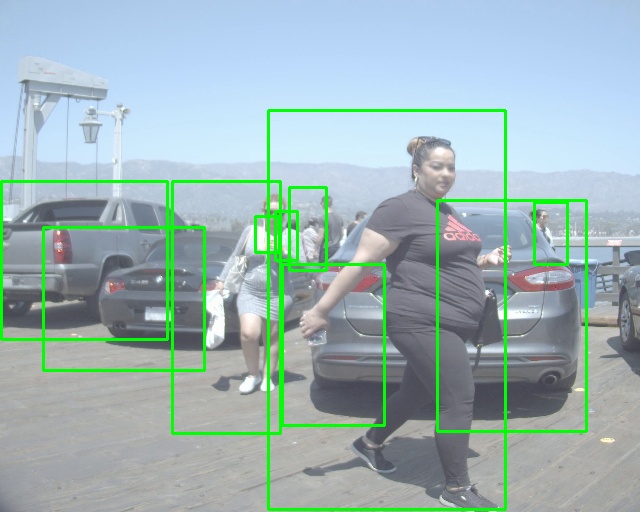}
  }
  \centerline{
      \includegraphics[width=0.22\linewidth]{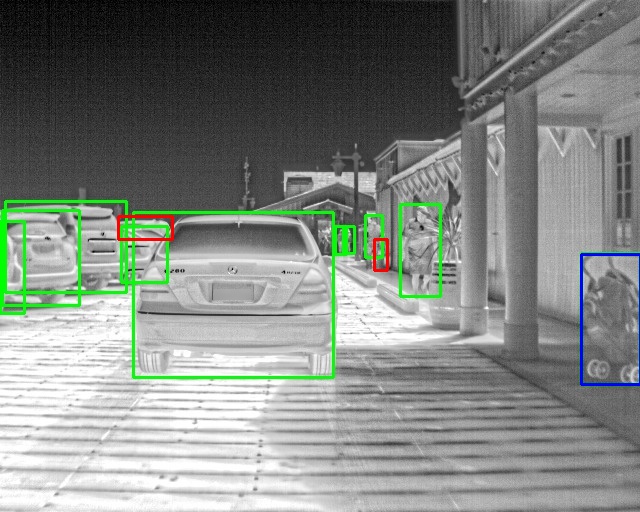}
      \includegraphics[width=0.22\linewidth]{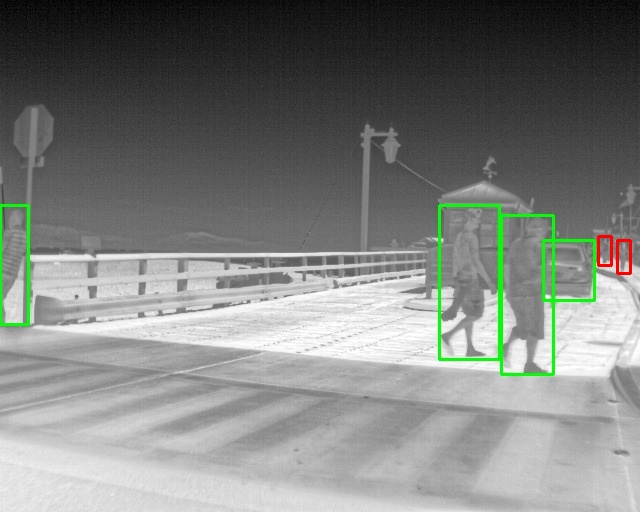}
      \includegraphics[width=0.22\linewidth]{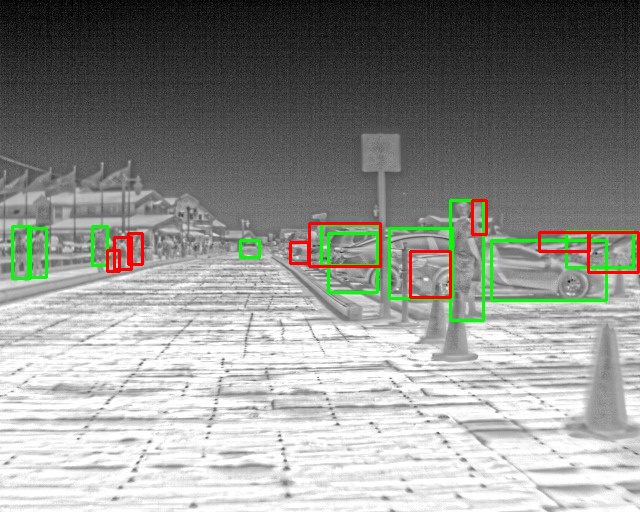}
      \includegraphics[width=0.22\linewidth]{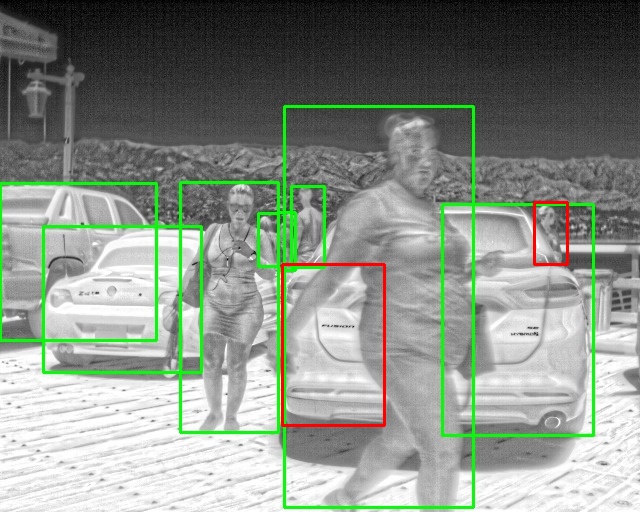}
  }
  \centerline{
      \includegraphics[width=0.22\linewidth]{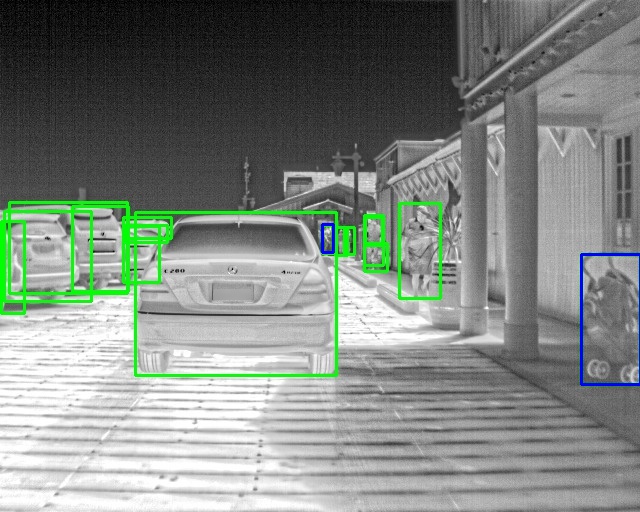}
      \includegraphics[width=0.22\linewidth]{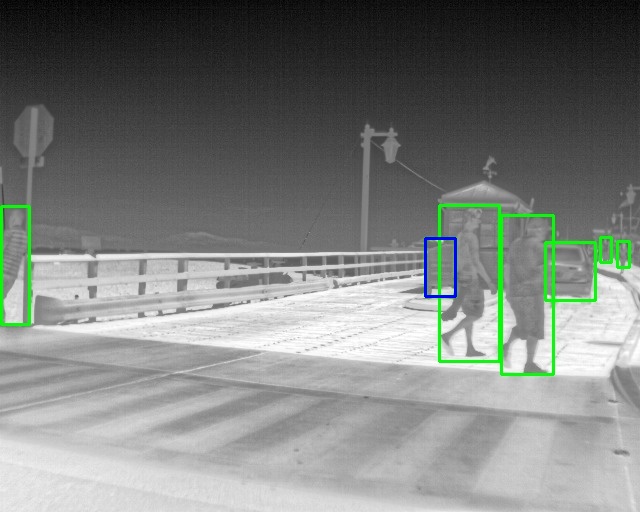}
      \includegraphics[width=0.22\linewidth]{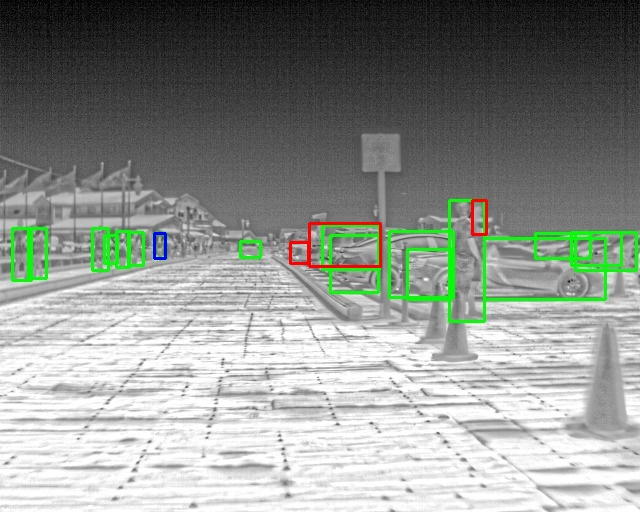}
      \includegraphics[width=0.22\linewidth]{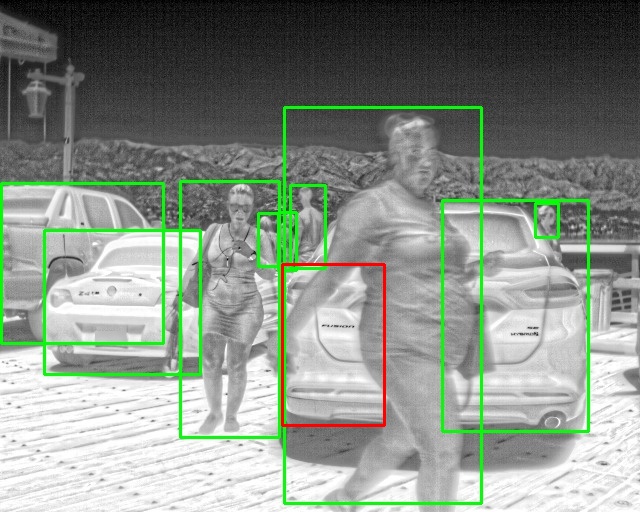}
  }
  \noindent\makebox[\linewidth]{\rule{0.8\paperwidth}{1pt}} 
  \vfill \bigskip
  \centerline{
      \includegraphics[width=0.22\linewidth]{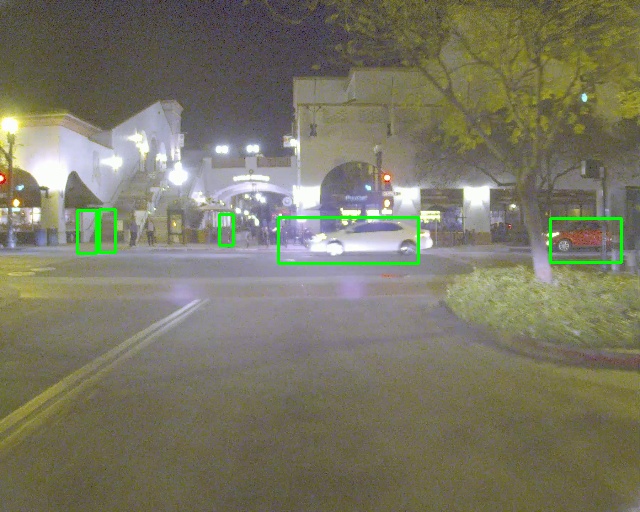}
      \includegraphics[width=0.22\linewidth]{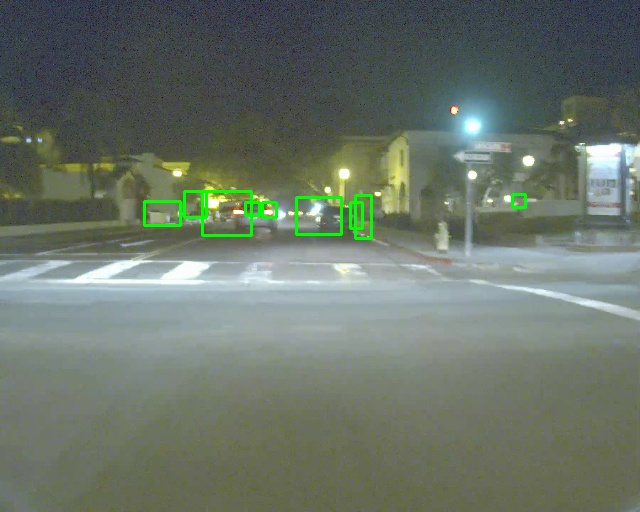}
      \includegraphics[width=0.22\linewidth]{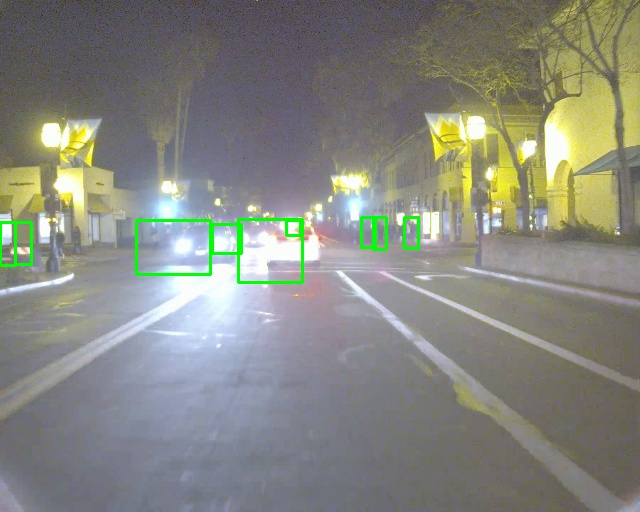}
      \includegraphics[width=0.22\linewidth]{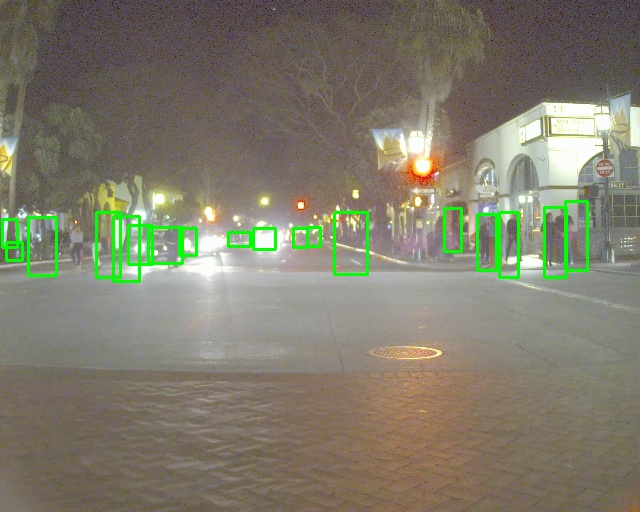}
  }
  \centerline{
      \includegraphics[width=0.22\linewidth]{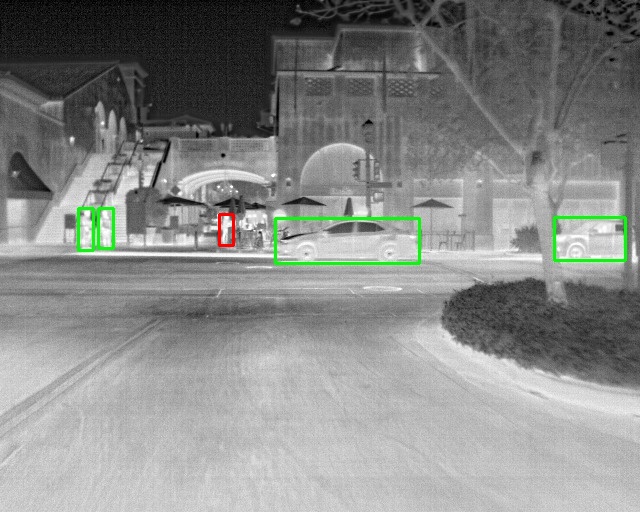}
      \includegraphics[width=0.22\linewidth]{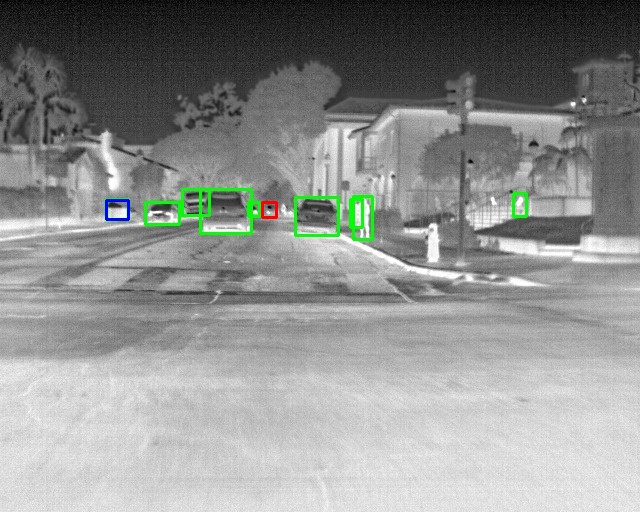}
      \includegraphics[width=0.22\linewidth]{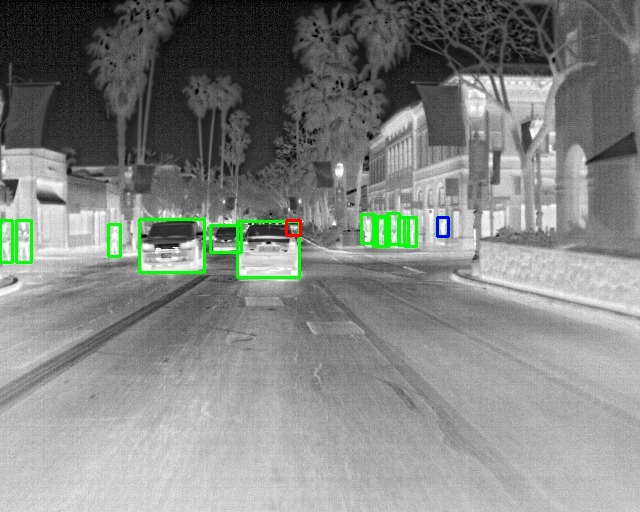}
      \includegraphics[width=0.22\linewidth]{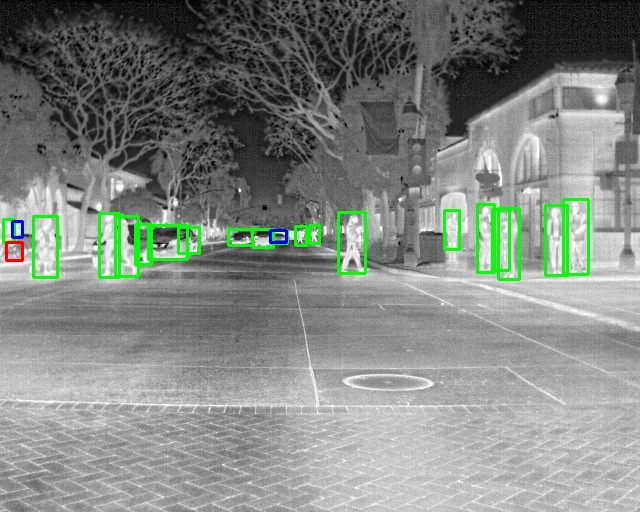}
  }
  \centerline{
      \includegraphics[width=0.22\linewidth]{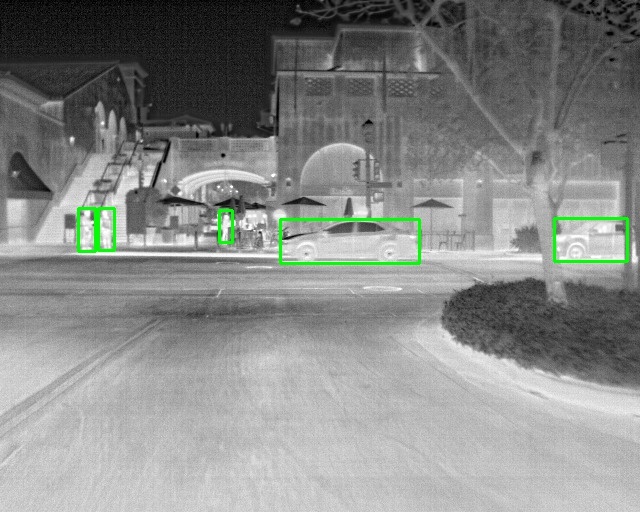}
      \includegraphics[width=0.22\linewidth]{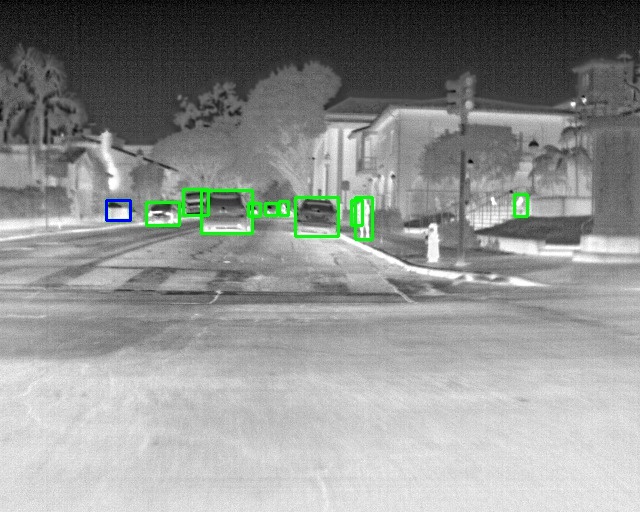}
      \includegraphics[width=0.22\linewidth]{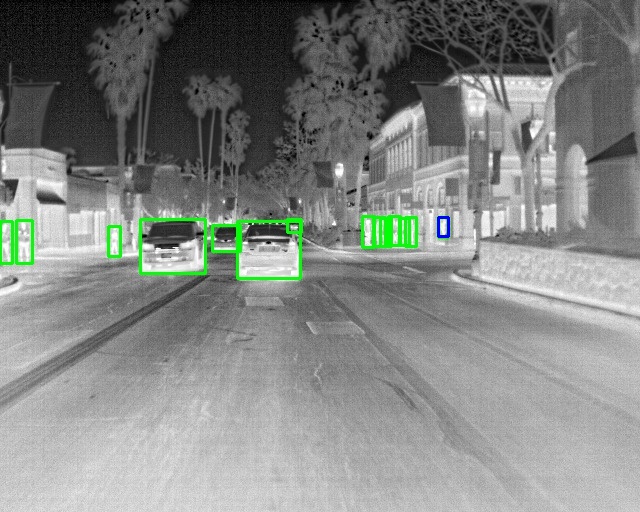}
      \includegraphics[width=0.22\linewidth]{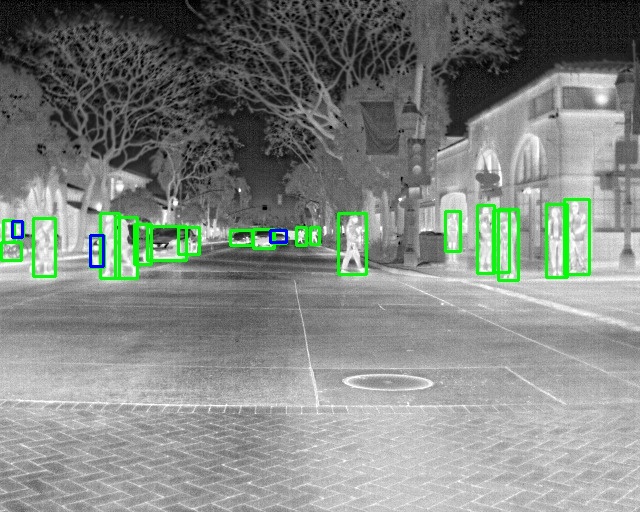}
  }
\vspace{-3mm}
\caption{ \small 
Qualitative results on more testing examples in FLIR dataset.
We place RGB-thermal images in triplet: in each macro row (divided by the black line), we show RGB images in the upper row and thermal images in two lower rows.
Over RGB images, we overlay ground-truth annotations,
highlighting that RGB and thermal images are strongly unaligned. To avoid clutter, we do not mark class labels for the bounding boxes.
On the thermal images, we show detection results from our thermal-only (mid-row) and best-performing ProbEn (with bounding box fusion) model  (bottom-row). 
{\color{darkgreen} Green}, {\color{red} red} and {\color{darkblue} blue} boxes stand for {\color{darkgreen} true positives}, {\color{red} false negative} (mis-detected persons) and {\color{darkblue} false positives}. 
}
\vspace{-3mm}
\label{fig:more-visual-FLIR}
\end{figure*}

Lastly, we provide more qualitative results in Figure~\ref{fig:more-visual-KAIST} and \ref{fig:more-visual-FLIR} for KAIST and FLIR, respectively.
Visually, we can see our ProbEn method performs  better than the compared methods.

\end{document}